\def\year{2021}\relax
\documentclass[letterpaper]{article} % DO NOT CHANGE THIS
\usepackage{aaai21}  % DO NOT CHANGE THIS
\usepackage{times}  % DO NOT CHANGE THIS
\usepackage{helvet} % DO NOT CHANGE THIS
\usepackage{courier}  % DO NOT CHANGE THIS
\usepackage[hyphens]{url}  % DO NOT CHANGE THIS
\usepackage{graphicx} % DO NOT CHANGE THIS
\usepackage{graphicx}  % DO NOT CHANGE THIS
\usepackage{natbib}  % DO NOT CHANGE THIS AND DO NOT ADD ANY OPTIONS TO IT
\usepackage{caption} % DO NOT CHANGE THIS AND DO NOT ADD ANY OPTIONS TO IT
\usepackage[switch]{lineno}
\usepackage{verbatim}
\usepackage{algorithm}
\usepackage{algpseudocode}
\usepackage{subfigure}
\usepackage{mathtools}
\usepackage{amsfonts}
\usepackage{booktabs}       % professional-quality tables
\usepackage{color}
\usepackage{hyperref}       % hyperlinks

\newtheorem{thm}{Theorem}[section]
\newtheorem{lem}{Lemma}[section]
\newtheorem{rem}{Remark}[section]

\title{Addressing Action Oscillations through Learning Policy Inertia}
\author{
    Chen Chen\textsuperscript{\rm 1}\thanks{Equal contributions. This work is done when Hongyao Tang is an intern at Noah's Ark Lab, Huawei.},
    Hongyao Tang\textsuperscript{\rm 2,1}\footnotemark[1],
    Jianye Hao\textsuperscript{\rm 1,2}\thanks{Corresponding author.},
    Wulong Liu\textsuperscript{\rm 1},
    Zhaopeng Meng\textsuperscript{\rm 2}

}

\affiliations{
    \textsuperscript{\rm 1}Noah's Ark Lab, Huawei \\
    \textsuperscript{\rm 2}College of Intelligence and Computing, Tianjin University

    chenchen9@huawei.com,
    bluecontra@tju.edu.cn, 
    \{haojianye,liuwulong\}@huawei.com,
    mengzp@tju.edu.cn

}

\begin{document}
%\linenumbers

\maketitle

\begin{abstract}
Deep reinforcement learning (DRL) algorithms have been demonstrated to be effective in a wide range of challenging decision making and control tasks.
However, these methods typically suffer from severe action oscillations in particular in discrete action setting, which means that agents select different actions within consecutive steps even though states only slightly differ. 
This issue is often neglected since the policy is usually evaluated by its cumulative rewards only. 
Action oscillation strongly affects the user experience and can even cause serious potential security menace
%due to the induced abnormal behavior,  
especially in real-world domains with the main concern of safety, such as autonomous driving.
To this end, we introduce Policy Inertia Controller (PIC) which serves as a generic plug-in framework to off-the-shelf DRL algorithms,
to enables adaptive trade-off between the optimality and smoothness of the learned policy in a formal way.
We propose Nested Policy Iteration as a general training algorithm for PIC-augmented policy which ensures monotonically non-decreasing updates under some mild conditions.
Further, we derive a practical DRL algorithm, namely Nested Soft Actor-Critic.
%, as a representative combination between PIC framework and Soft Actor-Critic.
Experiments on a collection of autonomous driving tasks and several Atari games suggest that our approach demonstrates substantial oscillation reduction in comparison to a range of commonly adopted baselines with almost no performance degradation.
\end{abstract}

\section{Introduction}
Deep reinforcement learning (DRL) has been widely considered to be a promising way to learn optimal policies in a wide range of practical decision making and control domains, 
such as Game Playing \cite{Mnih2015DQN,SilverHMGSDSAPL16AlphaGO}, Robotics Manipulation \cite{HafnerLB020Dream,Lillicrap2015DDPG,Smith19AVID},
%Autonomous Driving [+ref],
Medicine Discovery \cite{Popova19Molecule,schreck2019retrosyn,YouLYPL18GCPN} and so on.
One of the most appealing characteristics of DRL is that optimal policies can be learned in a model-free fashion,
even in complex environments with high-dimensional state and action space and stochastic transition dynamics.
%by model-free or model-based approaches without the access to exact environmental models,
%which are usually demanded in classic control approaches like Dynamic Programming \cite{Bertsekas09aNDP,SuttonB98RLAI}.

However, one important problematic phenomenon of DRL agent is \emph{action oscillation}, which means that a well-trained agent selects different actions within consecutive steps during online execution though the states only differ slightly, which leads to shaky behaviors and jerky trajectories.
Albeit the agent can achieve good task-specific rewards in simulation, 
the action oscillation may strongly affect the user experience in many practical interactive applications and exacerbate the wear and tear of 
%damage the motors on 
a real physical agent.
More crucially, the induced abnormal behavior can cause potential security menace in such as autonomous driving scenarios, where safety is the very first requirement.
In a nutshell, action oscillation inhibits the deployment of DRL agents in many real-world domains.

\begin{figure}
\centering
\hspace{-0cm}
\subfigure[\emph{Atari-Pong}]{
\includegraphics[width=0.21\textwidth]{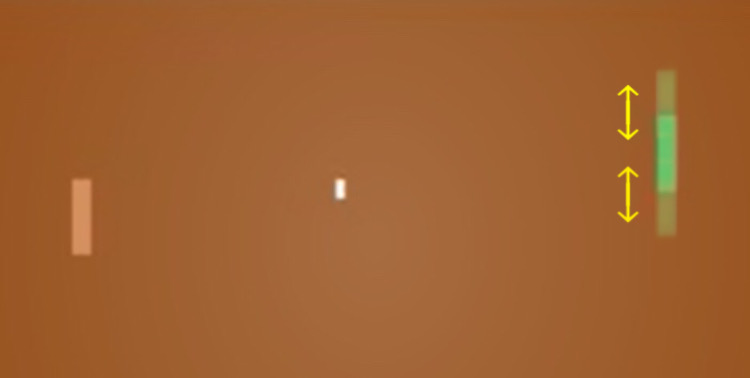}
\label{figure:general_policy}
}
\hspace{0.cm}
\subfigure[\emph{Highway-Overtaking}]{
\includegraphics[width=0.22\textwidth]{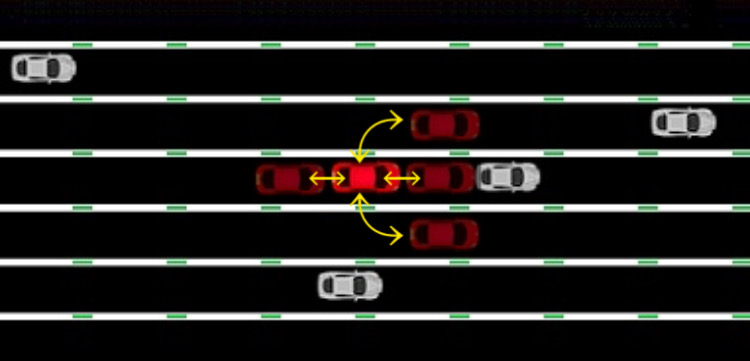}
\label{figure:action_oscillation}
}

\caption{Examples of action oscillation of DRL policies.
(a) In \emph{Atari-Pong}, a well-performing DQN agent often shows unnecessary up-down shakes in actions when controlling the bat; (b) In \emph{Highway-Overtaking}, a car agent learned by SAC can have frequent shifts between lane-change (i.e., left and right) and speed control (i.e., accelerate and decelerate) actions when driving on the lane.
%Frequent and unnecessary oscillations in actions is a usually-observed characteristic of learned DRL policies.
}
\label{figure:action_oscillation_example}
\end{figure}

Action oscillation can be widely observed for both deterministic policies and stochastic policies.
For deterministic policies like Deep Q-Network (DQN) \cite{Mnih2015DQN}, the underlying causes may come from the complexity of deep function approximation with high-dimensional inputs and stochastic noises due to partial observation and random sampling.
This issue can be more inevitable for stochastic policies.
For example, an entropy regularizer is often adopted in policy-based approaches; moreover, maximum entropy approaches, e.g., Soft Actor-Critic (SAC) \cite{HaarnojaZAL18SAC}, take policy entropy as part of optimization objective.
Such approaches encourage diverse behaviors of policy for better exploration and generalization, thus aggravate the oscillation in actions in turn.
Figure \ref{figure:action_oscillation_example} shows two exemplary scenarios in which `unnatural' and unnecessary oscillations in actions are often observed for learned policies.
It deserves to note that the 
action oscillation issue we study in this paper is different from the inefficient shaky or unstructured exploration behaviors studied in previous works \cite{sutton2011reinforcement,drive-smoothly-in-minutes,HaarnojaZAL18SAC,korenkevych2019autoregressive,haarnoja2018learning,kendall2019learning}.
On the contrary to exploration, we care about how to address action oscillation during online execution (i.e., exploitation). 

%In practice, the consequences of action oscillation are manifolds:
%on the one hand, it is so `unnatural'  when comparing with human beings making it discriminated from normal behaviors. 
%More crucially, it can cause potential security menace due to the induced abnormal behavior, which prevents DRL algorithms from being applied into domains with strong concerns of safety, such as autonomous driving scenarios.

%There exist some trick treatments that can be leveraged to reduce action oscillations. For instance, reward shaping can be done by adding action inconsistency penalty.
%However, it may bring oscillations reduction at the cost of violating the original reward function thus can be ineffective in challenging tasks. 

As previously explained, action oscillation is often neglected since mainstream evaluations of DRL algorithms are based on expected returns, which might be sufficient for games yet ignores some realistic factors in practical applications.
%Exploration is an integral part of RL that responsible for discovery of new behaviors. 
%It is typically achieved by executing a stochastic behavior policy \cite{sutton2011reinforcement} and such  unstructured exploration strategies  also tend to lead to shaky exploratory behavior and noisy, jittery trajectories \cite{drive-smoothly-in-minutes}. 
%Some solutions aim to address the inefficient and unsmoothed exploration by adopting  correlated noise \cite{HaarnojaZAL18SAC,korenkevych2019autoregressive}, low-pass filters \cite{haarnoja2018learning}, lower level controllers \cite{haarnoja2018learning,kendall2019learning}. 
%However, these methods all focus on the exploration phase rather than the execution phase and mainly apply in continuous action space setting. 
%For discrete action space setting, the 
We are aware that the idea of action repetition \cite{DurugkarRDM16Macro,Lakshminarayanan16bFrame,SharmaLR17Repeat,metelli2020control} that repeatedly executes a chosen action for a number of timesteps
to improve policy performance, can potentially be used to alleviate action oscillation. 
%Intuitively, these methods can potentially alleviate the action oscillation issue as well, even though they are originally targeted at performance improvement only.
%Besides, these works do not consider and evaluate the smoothness or action oscillation degree of learned policies.
To be specific,
%For example,
beyond the usual static frame skip mechanism, Lakshminarayanan et al. (\citeyear{Lakshminarayanan16bFrame}) propose dynamic frame skip to jointly learn dynamic repetition numbers through extending the action space with different repetition configurations.
Similarly in \cite{DurugkarRDM16Macro}, an action repetition policy is learned from the perspective of macro-actions that constitutes the same action repeated a number of times.
Further, Sharma et al. (\citeyear{SharmaLR17Repeat}) introduces FiGAR structure to address the scalability issue of above works by learning factored policies for actions and action repetitions separately.
Recently, Metelli et al. (\citeyear{metelli2020control}) analyze how the optimal policy can be affected by different fixed action repetition configurations, 
and present an algorithm to learn the optimal value function under given configurations. 

In general, action repetition utilizes temporal abstraction, which offers potential advantage in obtaining smooth policies.
However, temporal abstraction can decrease the sample efficiency since more simulation steps are needed to meet the same number of transition experiences when compared to a flat policy.
Besides, it is unclear whether the dynamic changes in underlying model (i.e., Semi-Markov Decision Process) will hamper the learning process.
From another angle, Shen et al. \cite{Shen20Smooth} propose to enforce smoothness in the learned policy with smoothness-inducing regularization which can  relieve the action oscillation problem in some degree. However, such constrains cannot guarantee the smoothness of one action sequence during execution, due to the essential difference of policy smoothness and action sequence smoothness.

In this paper, we propose Policy Inertia Controller (PIC) to address the action oscillation of DRL algorithms in discrete action space setting.
%and encourage the emergence of stable and smooth policies.
%PIC serves as a general framework that 
PIC serves as a general accessory to conventional DRL policies (called \emph{policy core} in the following) 
that adaptively controls the persistence of last action (i.e., \emph{Inertia})
%the smoothness of the original learned DRL policy,
according to the current state and the last action selected.
A PIC-augmented policy is built in a form of the linear combination between the distribution of the policy core and a Dirac distribution putting all mass on the last action.
Rather than introducing extra regularization and reward shaping in learning process, 
such a structure provides a more direct and fine-grained way of regulating the smoothness of the policy core at each time step.
%as well as a guarantee to reduce the action oscillation degree.
We also theoretically prove the existence of a family of smoother policies among PIC-augmented policies with equivalent or better performance than the policy core.

Further, we introduce Nested Policy Iteration (NPI) as a general training algorithm for PIC-augmented policy, consisting of an outer policy iteration process for the PIC module and an inner one for policy core that nested in the former.
We show that NPI can achieve monotonically non-decrease policy updates under some conditions.
Finally, we derive a practical DRL algorithm, namely Nested Soft Actor-Critic (NSAC), as a representative implementation of PIC framework with Soft Actor-Critic (SAC) algorithm \cite{HaarnojaZAL18SAC}.
%The PIC module is trained in the principle of Soft Policy Iteration \cite{HaarnojaZAL18SAC} within the scope of the mixed policy while keep the policy core fixed.
%The PIC module and the SAC policy core are trained simultaneously from online interactions.
%We show that the mixed policy induced by PIC-DSAC ensures monotonically non-decreasing updates under certain condition and maintains a guarantee of more smoothness in actions at the same time.
We demonstrate the effectiveness and superiority of our approach in addressing action oscillation among a variety of
%\footnote{Highway environments are originally provided at \url{https://github.com/eleurent/highway-env}.} 
autonomous driving environments in \emph{Highway} simulator and several Atari games in OpenAI Gym.
%in comparison with commonly adopted DRL baselines, as well as an ablation study to further understand our approach.
%}
%The results show a consistent superiority of our approach when compared with benchmark approaches in terms of action oscillation reduction while maintaining a high performance.

Our main contributions are summarized as follows:
\begin{itemize}
    \item 
    We propose Policy Inertia Controller (PIC) as a generic framework to address action oscillation issue, which is of significance to practical applications of DRL algorithms.
    
    \item 
    We propose a novel Nested Policy Iteration as a general training algorithm for the family of PIC-augmented policy, 
    as well as analyze the conditions when non-decrease policy improvement can be achieved.
    %which provably ensures non-decrease policy improvement.
    
    \item 
    Our extensive empirical results in a range of autonomous driving tasks and Atari game tasks show that our proposed approach can achieve substantial oscillation reduction at almost no performance degradation.
\end{itemize}

\section{Background}
\label{gen_inst}

%This section provides relevant background knowledge required in the remainder of the paper. 

\subsection{Markov Decision Process}
We consider a Markov Decision Process (MDP) $\mathcal{M} \coloneqq \langle \mathit{S}, \mathit{A}, r, \mathit{P}, \rho_{0}, \gamma, T \rangle$ where $\mathit{S}$ is the state space, $\mathit{A}$ is the
finite action space, $r:\mathit{S} \times \mathit{A} \rightarrow \mathbb{R}$ the bounded reward function, $\mathit{P}:S \times A \times S \rightarrow \mathbb{R}_{\in [0,1]}$ is the transition probability distribution, $\rho_0: \mathit{S} \rightarrow \mathbb{R}_{\in [0,1]}$ is the initial state distribution, $\gamma$ is the discount factor that we assume
$\gamma \in [0,1)$ and $T$ is the episode horizon.
The agent interacts with the MDP at discrete timesteps by performing its policy $\pi: S \times A \rightarrow \mathbb{R}_{\in [0,1]}$, 
generating a trajectory of states and actions,
$\tau = \left(s_0, a_0, s_1, a_1, \dots s_T, a_T \right)$, where $s_0 \sim \rho_0(s)$, $a_t \sim \pi(\cdot | s_t)$ and $s_{t+1} \sim \mathit{P}(\cdot |s_t, a_t)$.
The objective of a reinforcement learning agent is to 
find a policy that 
maximize the expected cumulative discounted reward:
\begin{equation}
\label{eqation:RL-objective}
\begin{aligned}
    %\pi^{*} = \mathop{\arg\max}_{\pi} \mathit{J}(\pi), \ \text{where} \ 
    \mathit{J}(\pi) = \mathbb{E}_{s_t, a_t \sim \rho_{\pi}} \left[\sum_{t=0}^{T}\gamma^{t} r(s_t, a_t) \right],
\end{aligned}
\end{equation}
where $\rho_{\pi}$ is the
% distribution of state-action pairs induced by policy $\pi$.
state-action marginals of the trajectory distribution induced by policy $\pi$.
Thus, the optimal policy is $\pi^{*} = \mathop{\arg\max}_{\pi} \mathit{J}(\pi)$.

In reinforcement learning,
the state-action value function $Q:S\times A \rightarrow \mathbb{R}$ is defined as the expected cumulative discounted reward for selecting action $a$ in state $s$, then following a policy $\pi$ afterwards:
$Q^{\pi}(s,a) = \mathbb{E}_{\pi} \left[\sum_{t=0}^{T}\gamma^{t} r(s_t, a_t)|s_0=s, a_0=a \right]$.
Similarly, the state value function $V$ denotes value under a certain state $s$, i.e., $V^{\pi}(s) = \mathbb{E}_{\pi} \left[\sum_{t=0}^{T}\gamma^{t} r(s_t, a_t)|s_0=s \right]$.

\subsection{Soft Actor-Critic}
Soft Actor-Critic (SAC) \cite{HaarnojaZAL18SAC} is an off-policy reinforcement learning algorithm that optimizes a stochastic policy that maximizes the maximum entropy objective:
\begin{eqnarray}
\label{MaxEnt_Objective}
    %\pi^{*} = \mathop{\arg\max}_{\pi} \sum_{t=0}^{T} \mathbb{E}_{(s_t, a_t) \sim \rho_{\pi}} \big[\gamma^{t}\big( r(s_t,a_t) +\alpha \mathcal{H}(\pi(\cdot|s_t)) \big) \big]
    \mathit{J}_{\text{Ent}}(\pi) = \mathbb{E}_{s_t, a_t \sim \rho_{\pi}} \left[ \sum_{t=0}^{T} \gamma^{t}\big( r(s_t,a_t) +\alpha \mathcal{H}(\pi(\cdot|s_t)) \big) \right]
\end{eqnarray}
%$\pi$ is a policy, $\pi^{*}$ is the optimal policy, 
% $T$ is the number of timesteps, $r:\mathit{S} \times \mathit{A} \rightarrow \mathbb{R} $ is the reward function, $\gamma \in [0,1]$ is the discount rate,
%$s_t \in \mathit{S}$ is the state at timestep $t$, $a_t \in \mathit{A}$ is the action at timestep $t$, 
%where $\rho_{\pi}$ is the distribution of state-action pairs induced by policy $\pi$,
where
the temperature parameter $\alpha$ determines the relative importance of the reward versus the policy entropy term at state $s_t$, $\mathcal{H}(\pi(\cdot|s_t)) = -\mathbb{E}_{a_t \sim \pi }\log \pi(\cdot |s_t)$.
%and $\mathcal{H}(\pi(\cdot|s_t))$ is the entropy of the policy $\pi$ at state $s_t$ and is calculated as $\mathcal{H}(\pi(\cdot|s_t)) = -\mathit{E}_{a_t \sim \pi(\cdot |s_t) }\log \pi(\cdot |s_t)$.

SAC uses \emph{soft policy iteration} which alternates between policy evaluation and policy improvement within the maximum entropy framework. 
The policy evaluation step involves computing the values of policy $\pi$ through repeatedly applying the modified Bellman backup operator $\mathit{T}^{\pi}$ as:
%To do this they first define the soft state value function as:
\begin{eqnarray}
\mathcal{T}^{\pi}Q(s_t,a_t) = r(s_t,a_t) + \gamma \mathbb{E}_{s_{t+1}\sim P}[V(s_{t+1})],
\label{soft_q_definition} \\
%\end{eqnarray}
%\begin{eqnarray}
\text{where} \ V(s_t) = \mathbb{E}_{a_t \sim \pi}\big[ Q(s_t,a_t)-\alpha\log(\pi(a_t|s_t))  \big], \nonumber
\label{soft_v_definition}
\end{eqnarray}
where $Q(s_t,a_t)$ and $V(s_t)$ here denote the soft variants of value functions.
%and $P$ represents the state distribution induced by $\pi$.
%Then they can obtain the soft state action-function repeatedly applying the modified Bellman backup operator $\mathit{T}^{\pi}$ given by
The policy improvement step then involves updating the policy towards the exponential of the soft $Q$-function, 
%which is a direction that maximizes the rewards it will achieve, 
%To make the policy tractable, it restricts the possible policies to a parameterised family of distributions, 
with the overall policy improvement step given by:
\begin{eqnarray}
\pi_{\rm new} = \mathop{\arg\min}_{\pi \in \Pi} D_{\rm KL}\bigg( \pi(\cdot|s_t) \| \frac{\exp\big(\frac{1}{\alpha}Q^{\pi_{\rm old}}(s_t,\cdot)\big)}{Z^{\pi_{\rm old}}(s_t)}  \bigg),
\label{soft_pi_definition}
\end{eqnarray}
where $\Pi$ denotes the policy space and the partition function $Z^{\pi_{\rm old}}(s_t)$ is intractable but does not contribute to the gradient with respect to the new policy thus can be ignored.

With continuous states, the soft $Q$-function $Q_{\theta}(s_t,a_t)$ is approximated with parameters $\theta$ via minimizing the soft Bellman residual according to (\ref{soft_q_definition}). 
The policy $\pi_{\phi}(a_t|s_t)$ that parameterized with parameters $\phi$ is learned by minimizing the expected KL-divergence (\ref{soft_pi_definition}).
%after multiplying by the temperature parameter $\alpha$ and ignoring the partition function $Z^{\pi_{\rm old}}(s_t)$.
In the original paper, the authors propose the practical algorithm in continuous action setting by applying the reparameterization trick to Gaussian policy $\pi$ and utilize an additional $V$-network to stable the training.
Two separate $Q$-networks are adopted and the minimum of them is used as $Q$ estimates to reduce the overestimation issue \cite{Hasselt10DoubleQ}.
The temperature parameter $\alpha$ can also be learned to automatically adapt through introducing a target entropy hyperparameter and optimizing the dual problem \cite{Haarnoja19SACapplication}.

The discrete action version of SAC is derived in \cite{Christodoulou19SACdiscrete}, where the policy network outputs a multinomial over finite actions rather than a Gaussian distribution so that $V$-function can be estimated directly and no long need Monte-Carlo estimates.
In this paper, we focus on discrete action setting and consider the discrete SAC as policy core by default.
For continuous action case, our approach can also be applied with a few modification which is beyond the scope of this paper and we leave it for future work.
\begin{figure}
	\centering
	\includegraphics[width=0.46\textwidth]{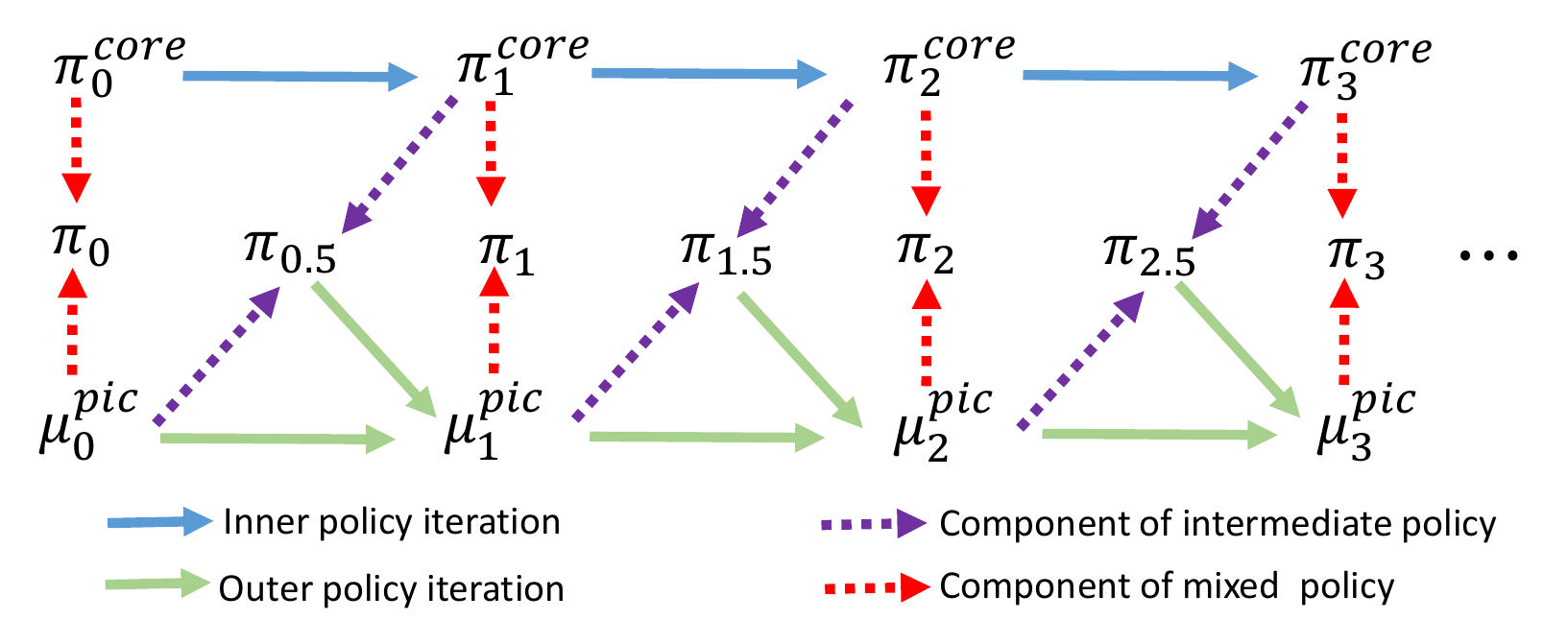} 
	\caption{Sketch map of Nested Policy Iteration, $\pi_{t}$ and $\pi^{\rm core}_{t}$ denote the mixed policy and policy core at $t$ respectively,  and $\pi_{t.5}$ denotes the intermediate status of  $\pi_t$ when policy core $\pi^{\rm core}_{t}$ has been updated in the inner policy iteration  yet $\mu^{\rm pic}$ has not. }
	\label{figure:alg_chart}
\end{figure}

\section{Approaches}

In this section, we first introduce the Policy Inertia Controller (PIC) framework,
then we propose Nested Policy Iteration (NPI) to train PIC and the policy core in a general way. 
Finally, we propose a practical algorithm, Nested Soft Actor-Critic with Policy Inertia Controller (PIC-NSAC), that combines PIC framework and SAC algorithm.

\subsection{Policy Inertia Controller Framework}
Before introducing Policy Inertia Controller (PIC), we first introduce a practical metric that measures the degree of action oscillation of a policy $\pi$ formally.
We define action oscillation ratio $\xi(\pi)$ below:
\begin{equation}
\label{eqation:action-oscillation-ratio}
    \xi(\pi) = \mathbb{E}_{\tau \sim \rho_{\pi}^{\tau}} \left[\frac{1}{T} \sum_{t=1}^{T} \left( 1 - \mathbb{I}_{\{a_{t-1}\}}(a_{t}) \right) \right],
\end{equation}
where $\rho_{\pi}^{\tau}$ is the distribution of state-action trajectory induced by policy $\pi$ and $\mathbb{I}_{\mathit{A}}(x)$ denotes the indicator function with set $A$.
Intuitively, $\xi(\pi)$ indicates how smooth the actions selected by policy $\pi$ are when acting in the environment.
The lower of $\xi(\pi)$ means the  smoother of $\pi$.
A high $\xi(\pi)$ means that the policy tends to select different actions within consecutive timesteps, i.e., more severe action oscillation.

One straightforward way to address action oscillation is to introduce reward shaping of adding action inconsistency penalty or use similar regularization according to Equation (\ref{eqation:action-oscillation-ratio}).
However, the drawbacks of such mechanisms are apparent: they alter the original learning objective and the hyperarameters involved need to be tuned for different tasks.
Moreover, such approaches have no guarantee on how the smoothness of policy learned will be.
To this end, we propose Policy Inertia Controller (PIC) that regulates a DRL policy distribution directly as follows:
\begin{equation}
\label{equation:mixed_policy}
\begin{aligned}
    \pi(\cdot |s_t,a_{t-1}) = 
    & \ \mu^{\rm pic}(s_t,a_{t-1})\delta(a_{t-1}) \\ 
    & + \left(1-\mu^{\rm pic}(s_t,a_{t-1})\right) \pi^{\rm core}(\cdot|s_t),
\end{aligned}
\end{equation}
%\begin{eqnarray}
%\label{equation:mixed_policy}
%&~& \pi(\cdot |s_t,a_{t-1}) = \mu^{\rm pic}(s_t,a_{t-1})\delta(a_{t-1}) \nonumber\\ 
%&+& (1-\mu^{\rm pic}(s_t,a_{t-1})) \pi^{\rm core}(\cdot|s_t),
%\end{eqnarray}
where $\delta(a_{t-1}): A \rightarrow \mathbb{R}_{\in [0,1]}^{|A|}$ denotes a discrete Dirac function that puts all probability mass on the action executed at timestep $t-1$, 
and $\pi^{\rm core}$ denotes the policy core that is trained as usual.
%and $\pi^{\rm core}(\cdot|s_t): A \rightarrow \mathbb{R}_{\in [0,1]}^{|A|}$ denotes the policy core (i.e., the original DRL policy) to be trained as usual.
The policy inertia controller $\mu^{\rm pic}(s_t,a_{t-1}):S \times A \rightarrow \mathbb{R}_{\in [0,1]}$ outputs a scalar weight and 
the final policy $\pi(\cdot |s_t,a_{t-1})$ is a linear combination of the Dirac and the policy core.
We also call the PIC-augmented policy $\pi(\cdot |s_t,a_{t-1})$ as \emph{mixed policy} in the following of this paper for distinction to policy core.
%that is weighted by $\mu^{\rm pic}(s_t,a_{t-1})$.
In another view, $\mu^{\rm pic}(s_t,a_{t-1})$ can be viewed as a 1-dimensional continuous policy that regulates the inertia of policy core (i.e., the persistence of last action) 
%determines the regulation of policy smoothness 
depending on current state and the last action.

Next, we show that the structure of the mixed policy has the appealing property to ensure the existence of 
a family of smoother policies with equivalent or better performance than the policy core
%a not higher action oscillation ratio with a non-lower expected cumulative discounted reward given the appropriate $\mu^{\rm pic}(s_t,a_{t-1})$ 
(Theorem \ref{thm:smoothness}) as below:
%Formally, we have,
\begin{thm}
\label{thm:smoothness}
Given any policy core $\pi^{\rm core}(\cdot|s)$, there exists some $\mu^{\rm pic}$ such that  $\xi(\pi) \le \xi(\pi^{\rm core})$ and $J({\pi}) \geq J(\pi^{\rm core})$, where $\pi$ is the corresponding mixed policy counterpart (Equation \ref{equation:mixed_policy}). 
\end{thm}
Detailed proof is provided in Supplementary Material A.1.
%\ref{proof:smoothness}.
Theorem \ref{thm:smoothness} implies that we can obtain a better policy in terms of both action oscillation rate and expected return through optimizing $\mu^{\rm pic}(s_t,a_{t-1})$.

The mixed policy $\pi$ defined in Equation (\ref{equation:mixed_policy}) is a nested policy in which the policy core $\pi^{\rm core}$ is nested. 
%Theorem \ref{thm:smoothness} has proved the existence of the $\mu^{\rm pic}$ and 
In next section we introduce a general algorithm to train the nested policies, i.e.,  $\pi^{\rm core}$ and $\mu^{\rm pic}$, together.

%\textcolor{red}{[To add a illustration of PIC framework in this section.]}

\begin{algorithm}[tb]
	\caption{Nested Policy Iteration (NPI) for PIC-augmented Policy}
	\textbf{Input:} Policy evaluation and policy improvement processes $\mathcal{E}_{\rm in}$, $\mathcal{I}_{\rm in}$ for mixed policy $\pi^{\rm core}$, and $\mathcal{E}_{\rm out}$, $\mathcal{I}_{\rm out}$ for mixed policy $\pi$ (Equation (\ref{equation:mixed_policy}))
	\begin{algorithmic}[1]
	    \State Initialize policy core $\pi^{\rm core}$ and its corresponding mixed policy $\pi$
	    \For{Outer policy iteration number $t_{\rm out}$}
	        \For{Inner policy iteration number $t_{\rm in}$}
	            \State Evaluate the values of $\pi^{\rm core}$ until convergence with $\mathcal{E}_{\rm in}$ 
	            %\hfill \textcolor{blue}{\# Inner policy evaluation}
	            \State Improve $\pi^{\rm core}$ according to $\mathcal{I}_{\rm in}$
	            %\hfill \textcolor{blue}{\# Inner policy improvement}
	        \EndFor
	        \State Evaluate the values of $\pi$ until convergence with $\mathcal{E}_{\rm out}$
	        %\hfill \textcolor{blue}{\# Outer policy evaluation}
	        \State Improve $\pi$ (i.e., update $\mu^{\rm pic}$) according to $\mathcal{I}_{\rm out}$ while keep $\pi^{\rm core}$ fixed
	        %\hfill \textcolor{blue}{\# Outer policy improvement}
	    \EndFor
%	    \COMMENT Outer policy evaluation process
	\end{algorithmic}
\label{alg:NPI}
\end{algorithm}

\subsection{Nested Policy Iteration for PIC-augmented Policy}\label{section:NPI}
In this section, we propose Nested Policy Iteration (NPI), a general training algorithm for the family of mixed policy defined in Equation (\ref{equation:mixed_policy}).
%, including the PIC module $\mu^{\rm pic}$ and the policy core $\pi_{\rm core}$.
As in Algorithm \ref{alg:NPI}, NPI consists of an outer policy iteration and an inner policy iteration which is nested in the former one.
The policy core $\pi^{\rm core}$ is trained as usual according to inner policy iteration.
The outer policy iteration is in the scope of the mixed policy $\pi$ yet only the PIC module $\mu^{\rm pic}$ is updated during outer policy improvement. The 
sketch map of NPI is shown in Figure \ref{figure:alg_chart}.
In the following, we show that our proposed NPI can achieve monotonically non-decreasing updates for the mixed policy $\pi$.
%with respect to the objective in Equation (\ref{MaxEnt_Objective}).
%, $Q_{\rm new}(s,a) \ge Q_{\rm old}(s,a)$ for all $(s,a) \in S \times A$.

First, we show that an improvement for policy core $\pi^{\rm core}$ during the inner policy iteration can induce an improvement for the mixed policy $\pi$ as well (Lemma \ref{lemma:intermidiate_policy_improvement}).
We use $Q^{\rm core}(s_t,a_t)$, $Q(s_t,a_{t-1},a_t)$ to denote the $Q$-functions of $\pi^{\rm core}(\cdot|s_t)$ and  $\pi(\cdot|s_t,a_{t-1})$ respectively, and also adopt  subscripts \emph{old} and \emph{new} to indicate $Q$-functions and policies before and after one policy improvement iteration.
Specially, we further use subscript \emph{mid} to denote the intermediate status of the mixed policy $\pi$ when $\pi^{\rm core}$ has been updated in the inner iteration yet $\mu^{\rm pic}$ has not, i.e., $\pi_{\rm mid}(\cdot|s_t,a_{t-1}) = \mu_{\rm old}^{\rm pic}\delta(a_{t-1})+ \pi_{\rm new}^{\rm core}(\cdot|s_t)$.
Now we formalize above result in the following lemma.
%Consider the update process of how mixed policy $\pi$ is updated.
\begin{lem}
\label{lemma:intermidiate_policy_improvement}
\text{(Intermediate Policy Improvement).}
Given an inner policy iteration with policy improvement (i.e., $\pi^{\rm core}_{\rm old} \rightarrow \pi^{\rm core}_{\rm new}$ or $\pi_{\rm old} \rightarrow \pi_{\rm mid}$)  that ensures $Q^{\rm core}_{\rm new}(s_t,a_t) \ge Q^{\rm core}_{\rm old}(s_t,a_t)$ for all $(s_t,a_t)$,  
and assume  $\mu_{\rm old}^{\rm pic}(s_t,a_{t-1})$ satisfies the following inequality, 
\begin{eqnarray}
\label{equation:pic_ineq}
\mu_{\rm old}^{\rm pic}(s_t,a_{t-1}) 
\le \frac{\min_{(s_t,a_t)} (Q^{\rm core}_{\rm new}(s_t,a_t) - Q^{\rm core}_{\rm old}(s_t,a_t))}{N  \cdot C_0(\sum_{t}^{T}t\gamma^{t})} \nonumber
\end{eqnarray}
for all $(s_t,a_t,a_{t-1})$,
where $N\geq 4$ and $C_0$ is the upper bound of both $A^{\pi_{\rm new}^{\rm core}}$ and $A^{\pi_{\rm old}}$,
 then we have
\begin{eqnarray}
&~&Q_{\rm mid}(s_t,a_{t-1},a_t) -  Q_{\rm old}(s_t,a_{t-1},a_t)\nonumber \\
&\geq& (1-\frac{4}{N} )\min_{(s_t,a_t)} (Q^{\rm core}_{\rm new}(s_t,a_t) - Q^{\rm core}_{\rm old}(s_t,a_t)) \nonumber
\end{eqnarray}
for all $(s_t,a_{t-1},a_t) $ tuples.

\end{lem}
Detailed proof is provided in Supplementary Material A.2.
%\ref{proof:pic-improvement}.
\begin{rem}
Lemma \ref{lemma:intermidiate_policy_improvement} implies that 
the improvement of  policy core by inner policy iteration can also bring an improvement to the mixed policy given that $\mu^{\rm pic}_{\rm old} $ is appropriately small. 
Besides, the upper bound of tolerated $\mu^{\rm pic}_{\rm old} $ to guarantee such improvement 
and the increment from $Q_{\rm old}$ to $Q_{\rm mid}$ are both linearly dependent on the increment of policy core,
which means that bigger increment  in the inner iteration can tolerate bigger policy inertia $\mu^{\rm pic}_{\rm old}$,
and the increment for the intermediate mixed policy is proportional to the increment of policy core.
\end{rem}

\begin{figure*}
\centering
\hspace{-0.25cm}
\subfigure[Lane Change]{
\includegraphics[width=0.25\textwidth]{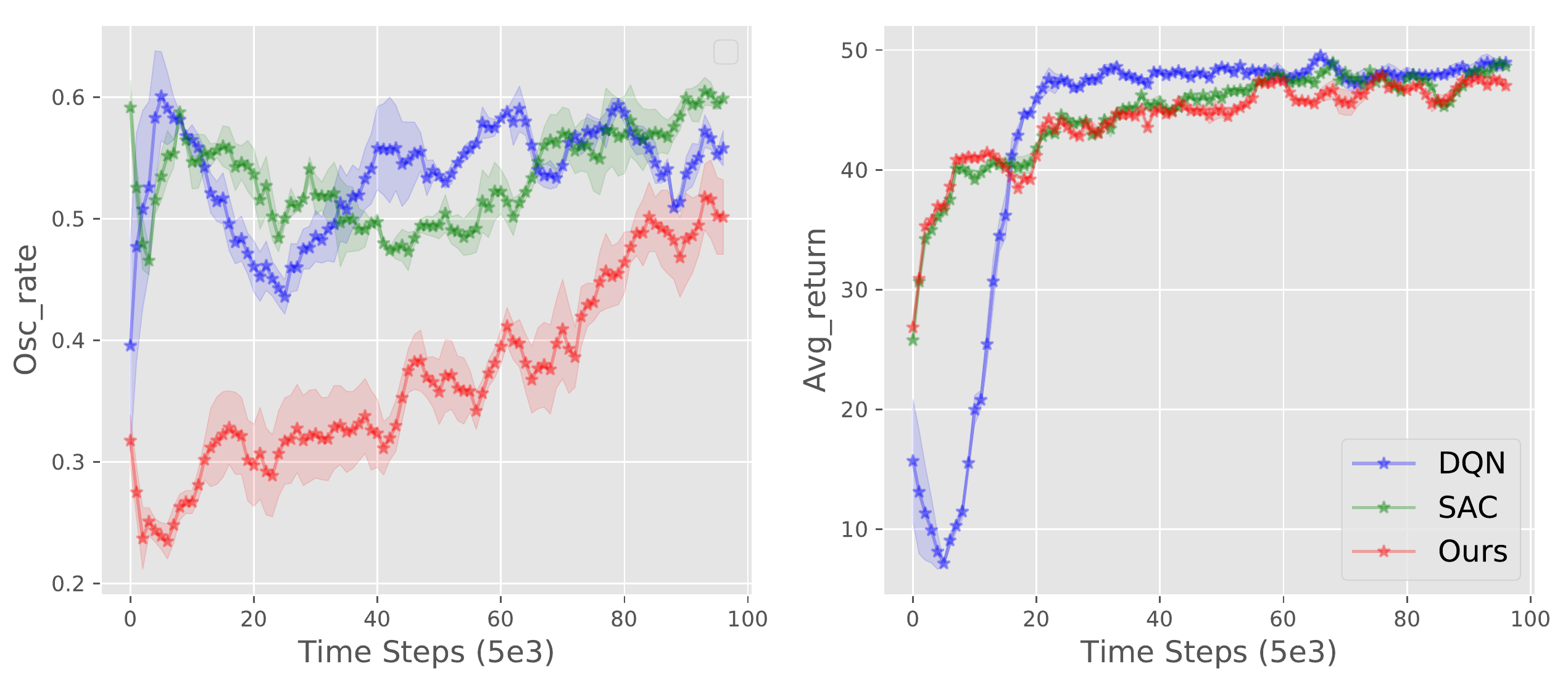}
}
\hspace{-0.35cm}
\subfigure[Merge]{
\includegraphics[width=0.25\textwidth]{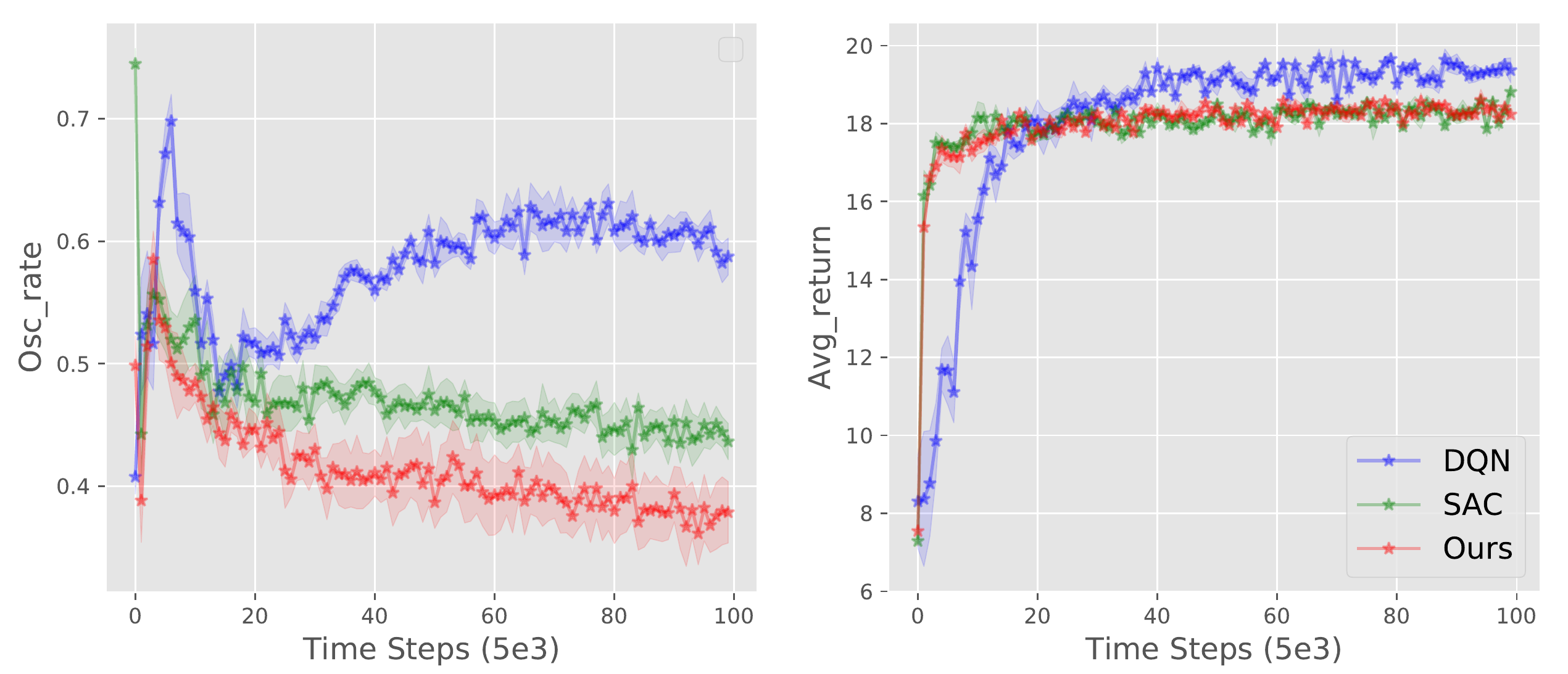}
}
\hspace{-0.35cm}
\subfigure[Intersection]{
\includegraphics[width=0.25\textwidth]{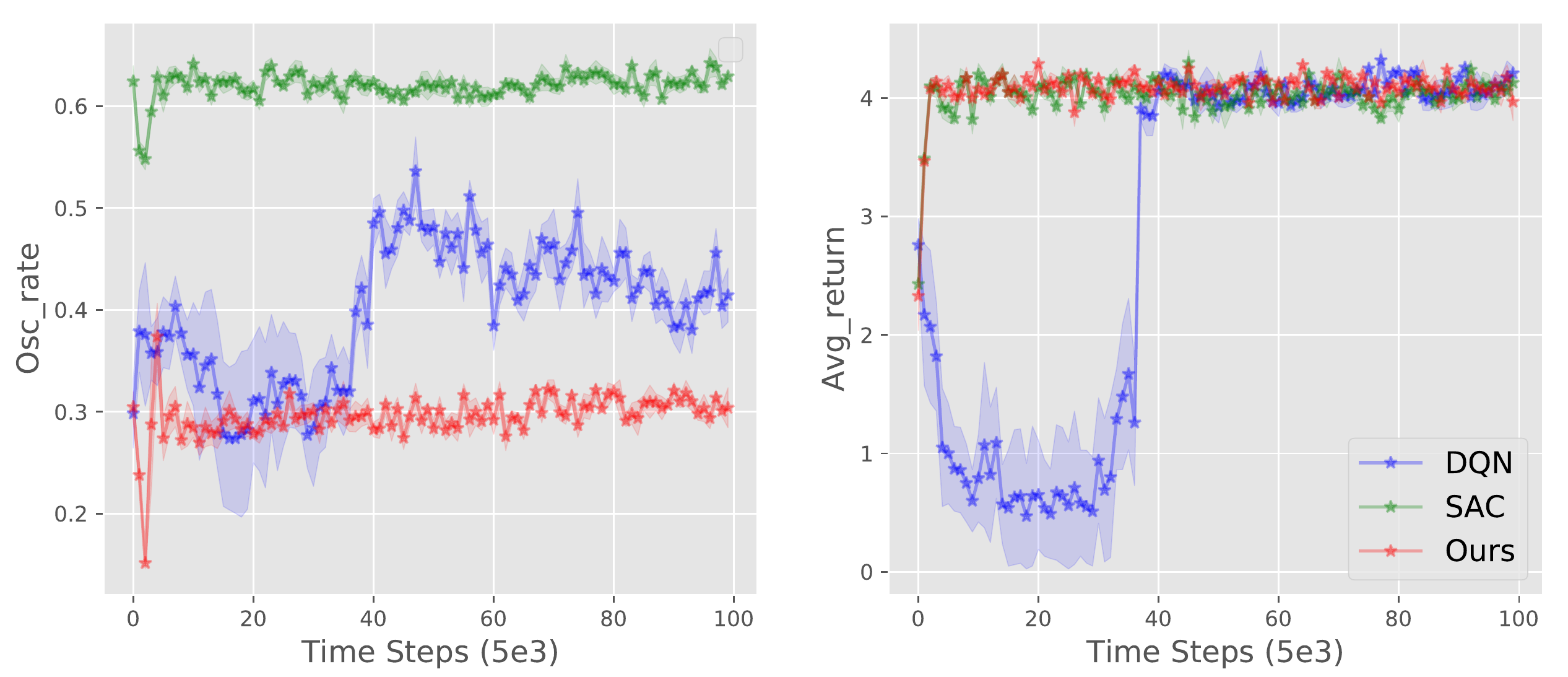}
}
\hspace{-0.35cm}
\subfigure[Two-way]{
\includegraphics[width=0.25\textwidth]{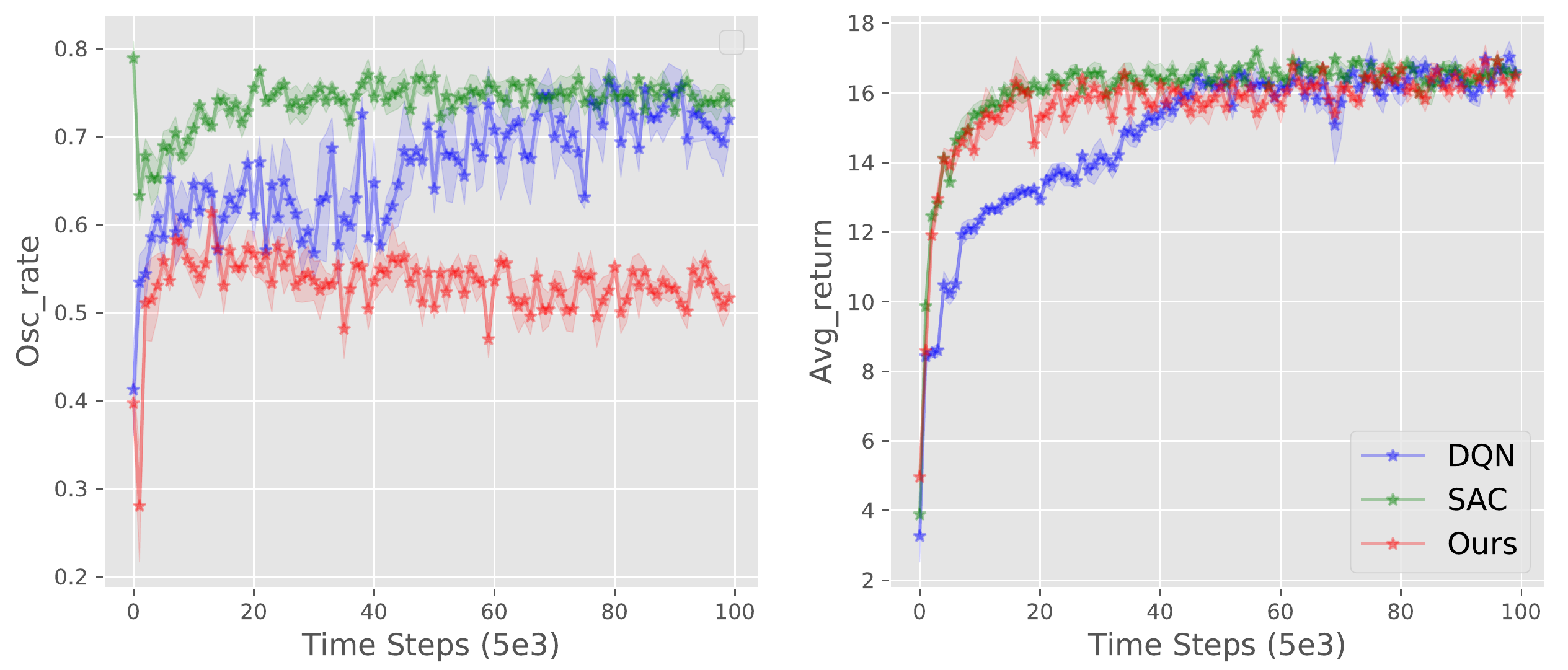}
}
\hspace{-0.5cm}
\subfigure[SpaceInvaders]{
\includegraphics[width=0.25\textwidth]{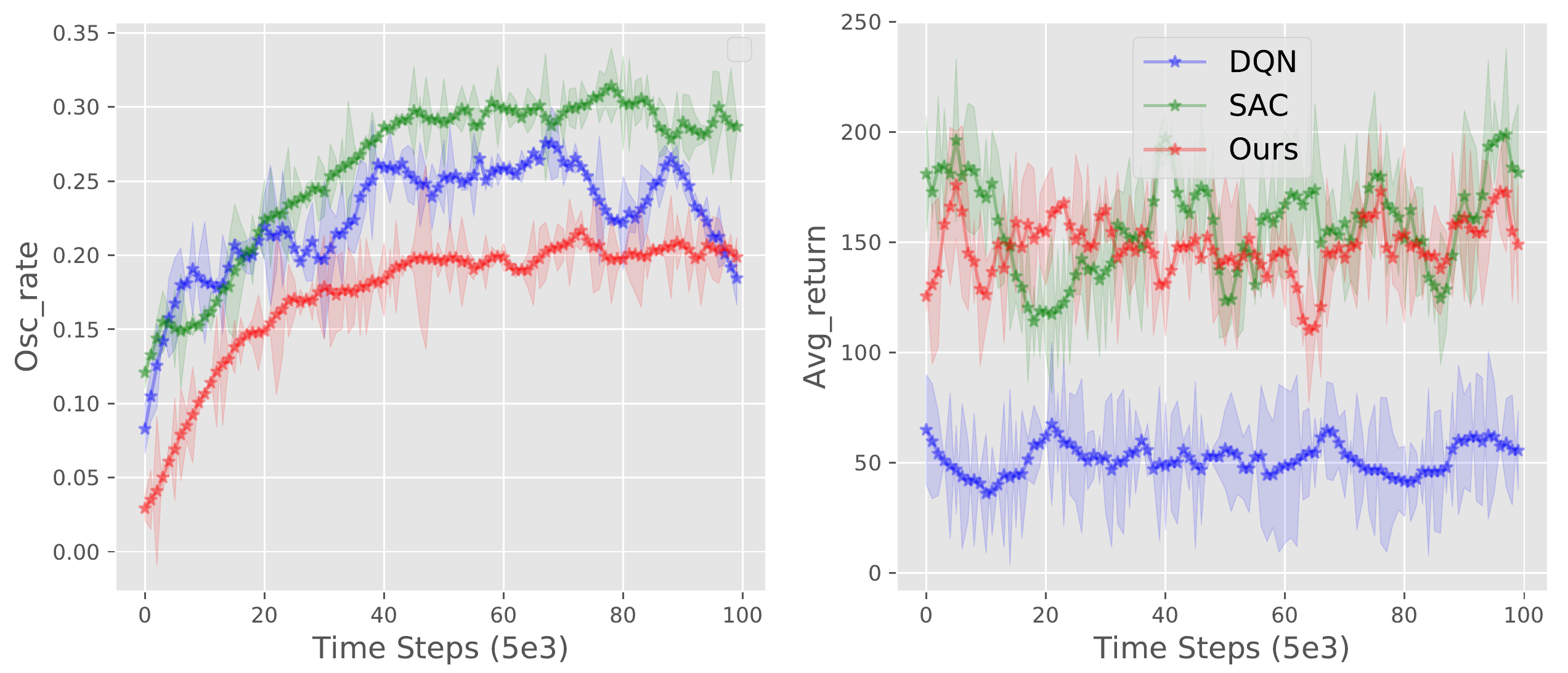}
}
\hspace{-0.4cm}
\subfigure[MsPacman]{
\includegraphics[width=0.25\textwidth]{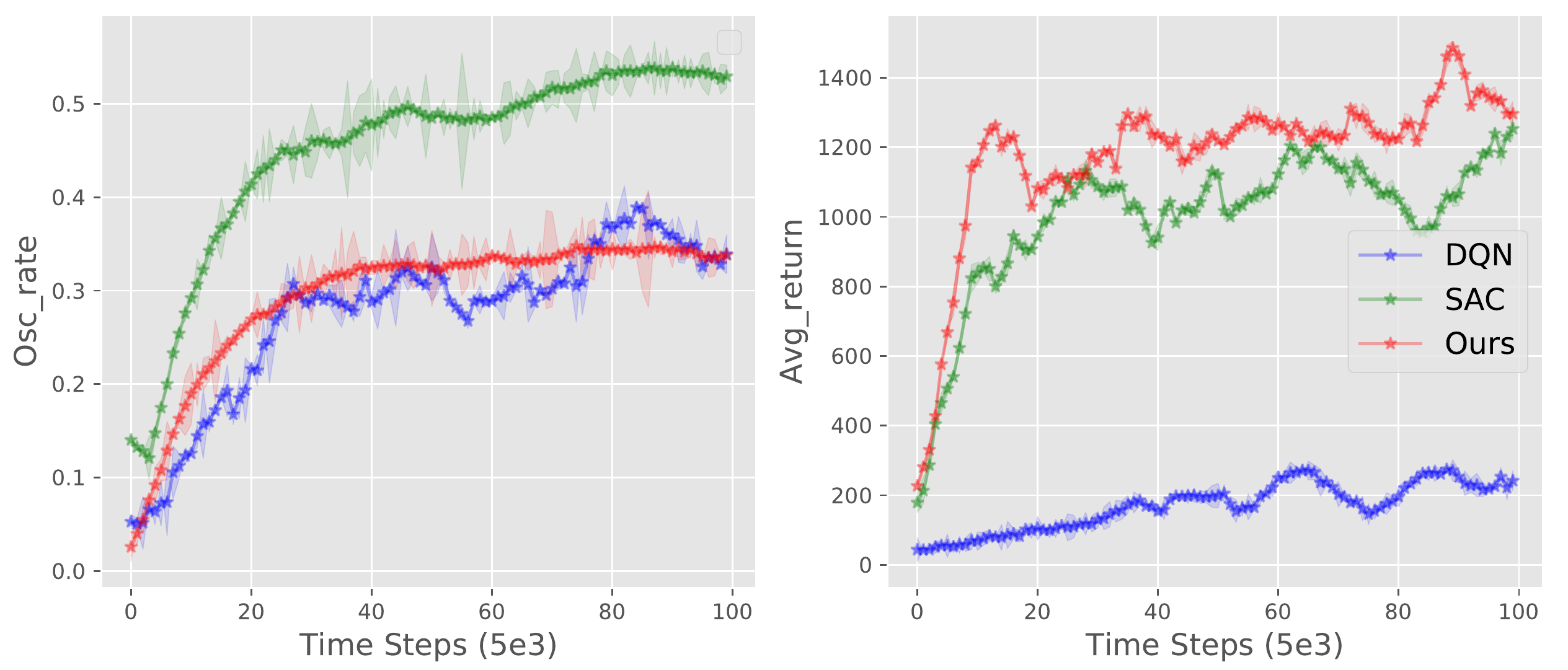}
}
\hspace{-0.4cm}
\subfigure[Qbert]{
\includegraphics[width=0.25\textwidth]{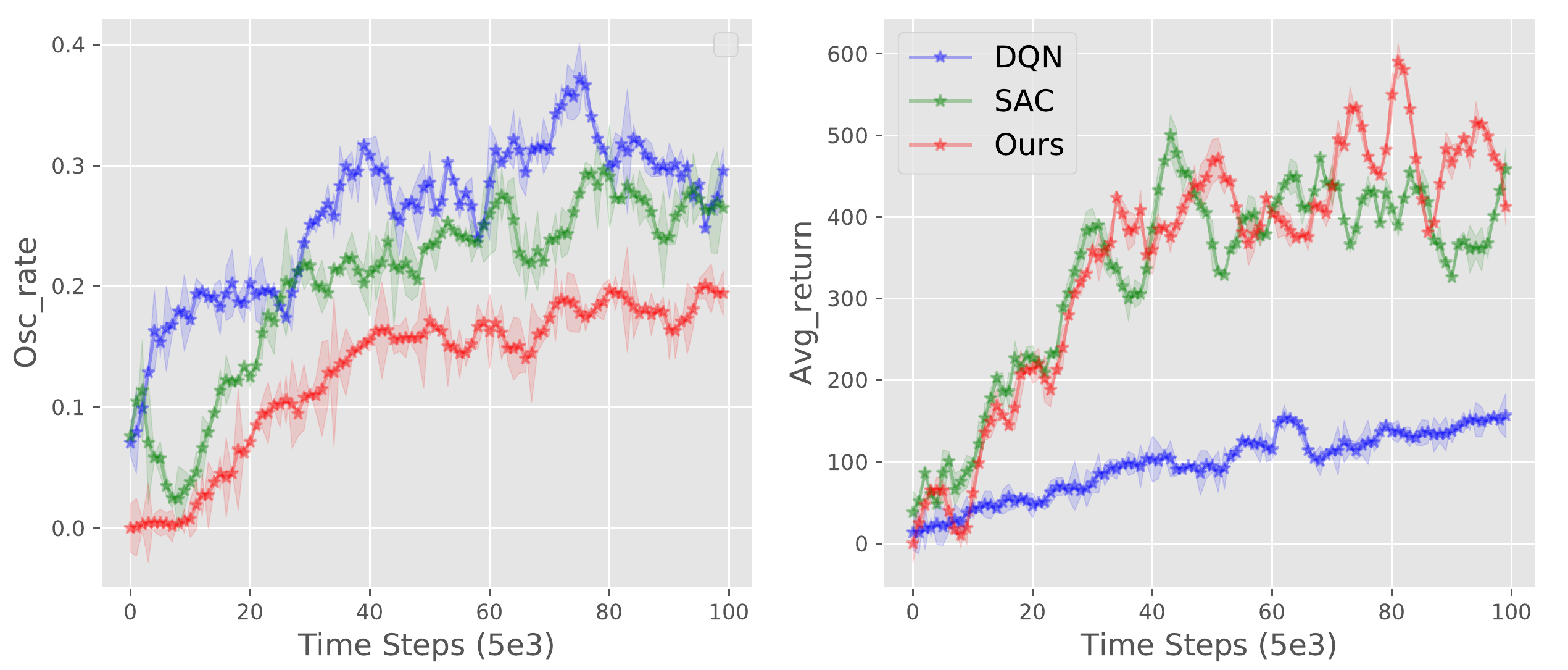}
}
\hspace{-0.4cm}
\subfigure[JamesBond]{
\includegraphics[width=0.25\textwidth]{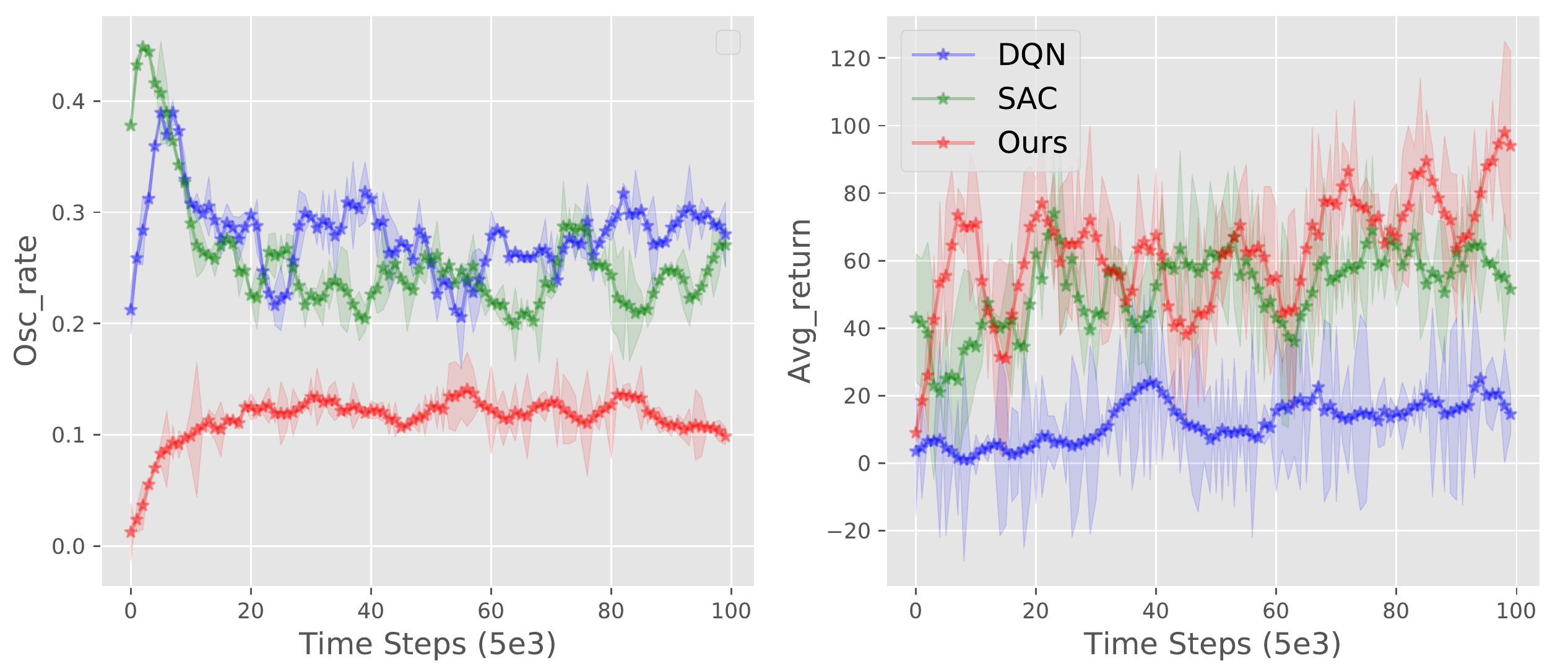}
}
\caption{Training curves of algorithms on \emph{(a)-(d)} four \emph{Highway} tasks and \emph{(e)-(h)} four Atari games w.r.t. action oscillation ratio (left) and returns (right). 
Our approach (red) consistently achieves lower action oscillation rate while retaining comparable performance across all tasks.
The horizontal axis denotes time step.
Results are means and one stds over 5 random seeds.}
\label{figure:base_results}
\end{figure*}

Next, 
%with a further improvement for the PIC module $\mu^{\rm pic}$ during the outer policy iteration, 
the full NPI algorithm alternates between
the inner policy iteration and the outer policy iteration
steps,
and it provably leads to a nested policy improvement (Lemma \ref{lemma:nested_policy_improvement}) and then monotonically non-decreasing updates for the mixed policy $\pi$ during overall NPI process (Theorem \ref{thm:NPI}).
\begin{lem}
\label{lemma:nested_policy_improvement}
\text{(Nested Policy Improvement).}
Based on Lemma \ref{lemma:intermidiate_policy_improvement}, given an outer policy iteration with policy improvement (i.e., $\mu^{\rm pic}_{\rm old} \rightarrow \mu^{\rm pic}_{\rm new}$ or $\pi_{\rm mid} \rightarrow \pi_{new}$) that ensures $Q_{\rm new}(s_t, a_{t-1}, a_t) \ge Q_{\rm mid}(s_t, a_{t-1}, a_t)$ for all $(s_t,a_{t-1},a_t)$ 
$\in S \times A \times A$,
%with $|A| < \infty$, 
then we have
$Q_{\rm new}(s_t,a_{t-1},a_t) \geq Q_{\rm old}(s_t,a_{t-1},a_t)$.
%for all $(s_t,a_{t-1},a_t) \in S \times A \times A$.
% \begin{eqnarray}
% Q_{\rm mid}(s_t,a_t) \geq Q_{\rm old}(s_t,a_t).
% \end{eqnarray} 
\end{lem}

Proof for Lemma \ref{lemma:nested_policy_improvement} can be easily obtained by chaining the inequalities between $Q_{\rm new}$, $Q_{\rm mid}$ and $Q_{\rm old}$,
which can be viewed as a two-step improvement obtained by inner and outer policy iteration respectively.

\begin{thm}
\label{thm:NPI}
\text{(Nested Policy Iteration).}
By repeatedly applying Lemma \ref{lemma:nested_policy_improvement},
the mixed policy $\pi$ achieves monotonically non-decreasing updates. 
\end{thm}
%Proof for Lemma \ref{lemma:nested_policy_improvement} can be easily obtained by chaining the inequalities between $Q_{\rm new}$, $Q_{\rm mid}$ and $Q_{\rm old}$, and 
Proof for Theorem \ref{thm:NPI} is a straightforward extension of Lemma \ref{lemma:nested_policy_improvement} in an iterative fashion under mild conditions.
%, as long as the two have same learning objective. 

According to \emph{Generalized Policy Iteration} (GPI) \cite{Sutton1988ReinforcementLA}, almost any RL algorithm can be interpreted in a policy iteration fashion.
Therefore, NPI can be viewed as a special case of GPI that combines any two RL algorithms for $\mu^{\rm pic}$ and $\pi^{\rm core}$ respectively, since no assumption is made on the choice of both the outer and the inner policy iteration.
%\textcolor{green}{[TO-CHECK]}
To make above theoretical results hold, the inner and the outer policy iteration should have the same learning objective, e.g., both common RL objective (Equation (\ref{eqation:RL-objective})) or maximum entropy objective (Equation (\ref{MaxEnt_Objective})).

\begin{rem}
The outer policy iteration of NPI algorithm is in the scope of the mixed policy $\pi$ that consists of $\mu^{\rm pic}$ and $\pi^{\rm core}$.
Since the PIC module $\mu^{\rm pic}(s_t,a_{t-1})$ can also be viewed as a continuous policy, one may conduct the outer policy iteration (i.e., an RL algorithm) in the scope of $\mu^{\rm pic}$ solely (instead of the mixed policy $\pi$).
However, such approach can be flawed in two aspects: 
first, the update signal can be very weak and stochastic since $\mu^{\rm pic}$ only indirectly influences the action selected;
moreover, the time-variant updates of the policy core are not considered explicitly during the evaluation process, thus resulting in a non-stationary environment for the learning of $\mu^{\rm pic}$.
%Therefore, the outer policy iteration of NPI is in the scope of the mixed policy $\pi$ rather than of $\mu^{\rm pic}$ only, since the information of the policy core can be reflected in such a way.
\end{rem}

\subsection{Nested Soft Actor-Critic}
%\subsection{Nested Soft Actor-Critic with Policy Inertia Controller}
\label{Method}
Based on the general NPI algorithm presented in previous section, we further derive a practical implementation with function approximation for continuous state space domains.
To be specific, we propose Nested Soft Actor-Critic with Policy Inertia Controller (PIC-NSAC) to train the PIC module $\mu^{\rm pic}$ and the policy core $\pi^{\rm core}$ simultaneously.

For the inner policy iteration, we resort to SAC algorithm because it is a representative of maximum entropy RL approaches that may typically suffer from action oscillation issue as we mentioned before.
We parameterize the policy core $\pi^{\rm core}_{\phi}(\cdot|s_t)$ with parameter $\phi$
%and its $Q$-function as $\pi^{\rm core}_{\phi}(\cdot|s_t)$ and $Q^{\rm core}_{\theta}(s_t,a_t)$ respectively, 
and update parameters $\phi$ and $\theta$ by SAC algorithm as usual.
As to the outer policy iteration, we also use SAC algorithm in the scope of the mixed policy to optimize the same objective as the inner one.
We approximate PIC module as $\mu^{\rm pic}_{\varphi}(s_t, a_{t-1})$ with parameter $\varphi$, thus obtain parameterized mixed policy $\pi_{\phi,\varphi}$.
The value networks of $\pi_{\phi,\varphi}$ is approximated during soft policy evaluation (Equation (\ref{soft_q_definition}))
%The $Q$-network $Q_{\varphi}(s_t, a_{t-1}, a_t)$ of $\pi_{\phi,\varphi}$ is approximated 
and only parameter $\varphi$ is updated during soft policy improvement (Equation (\ref{soft_pi_definition})).
All data used above comes from a replay buffer of past experiences collected using the mixed policy $\pi_{\phi,\varphi}$.

The overall algorithm (Algorithm 2)
%\ref{alg})
and complete formulations are provided in Supplementary Material B.
%\ref{section:overall_alg}.

%To avoid the extreme case that $\pi(\cdot|s_t,a_{t-1})$ degenerates into $\pi^{\rm core}(\cdot|s_t)$, we also add a constraint $\mu^{\rm pic}(s_t,a_{t-1}) \geq \xi_0$ with $\xi_0$ as a hyperparameter.

\begin{figure*}
\centering
\hspace{-0.3cm}
\subfigure[Intersection]{
\label{figure:repeat_intersection}
\includegraphics[width=0.33\textwidth]{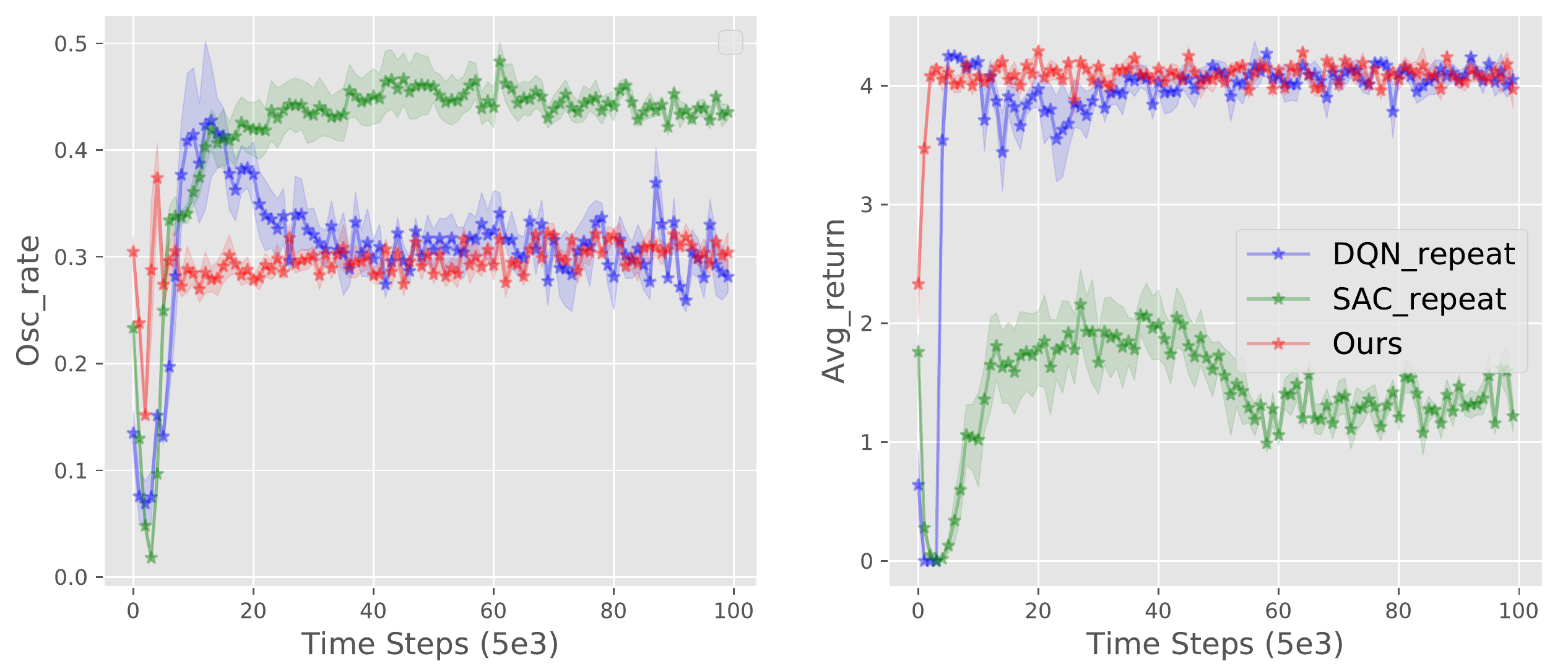}
}
\hspace{-0.3cm}
\subfigure[Two-Way]{
\label{figure:repeat_two-way}
\includegraphics[width=0.33\textwidth]{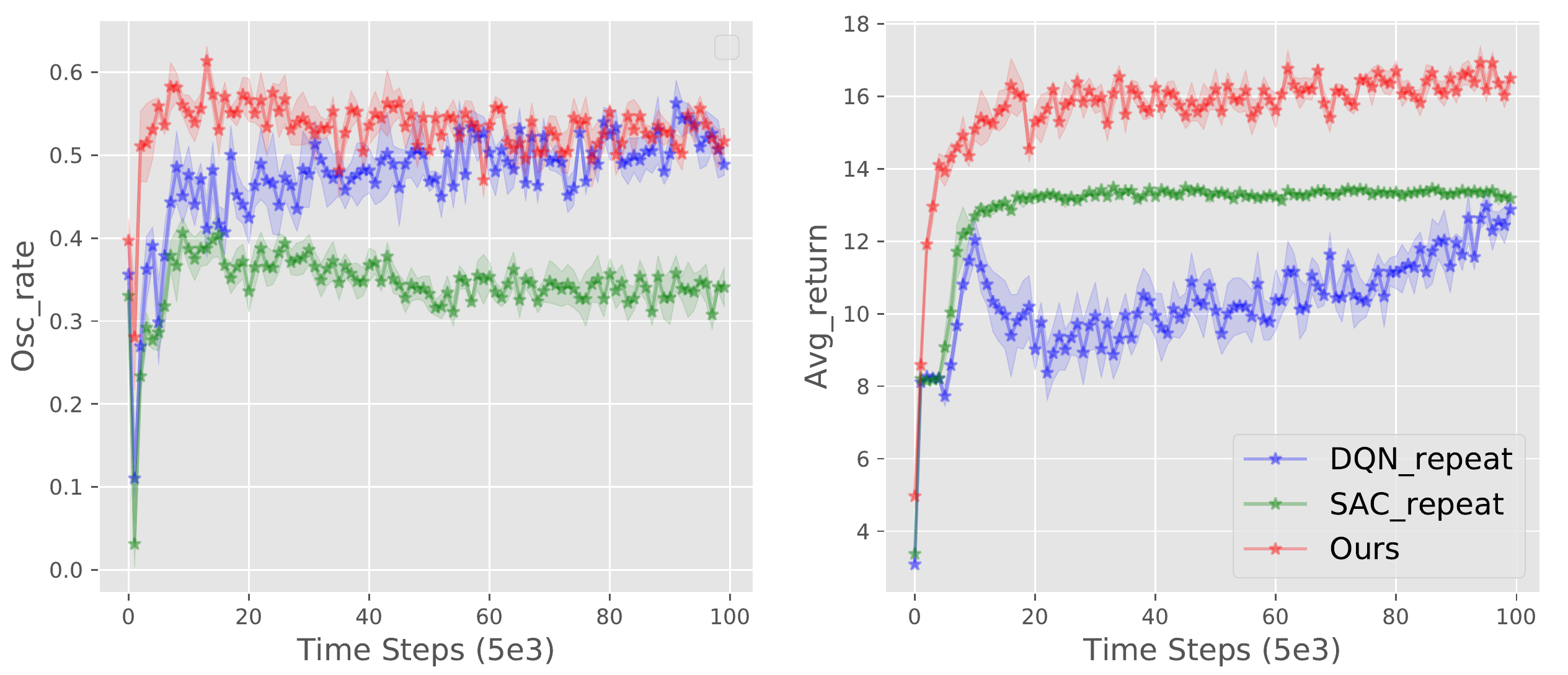}
}
\hspace{-0.3cm}
\subfigure[Intersection]{
\label{figure:penalty_intersection}
\includegraphics[width=0.33\textwidth]{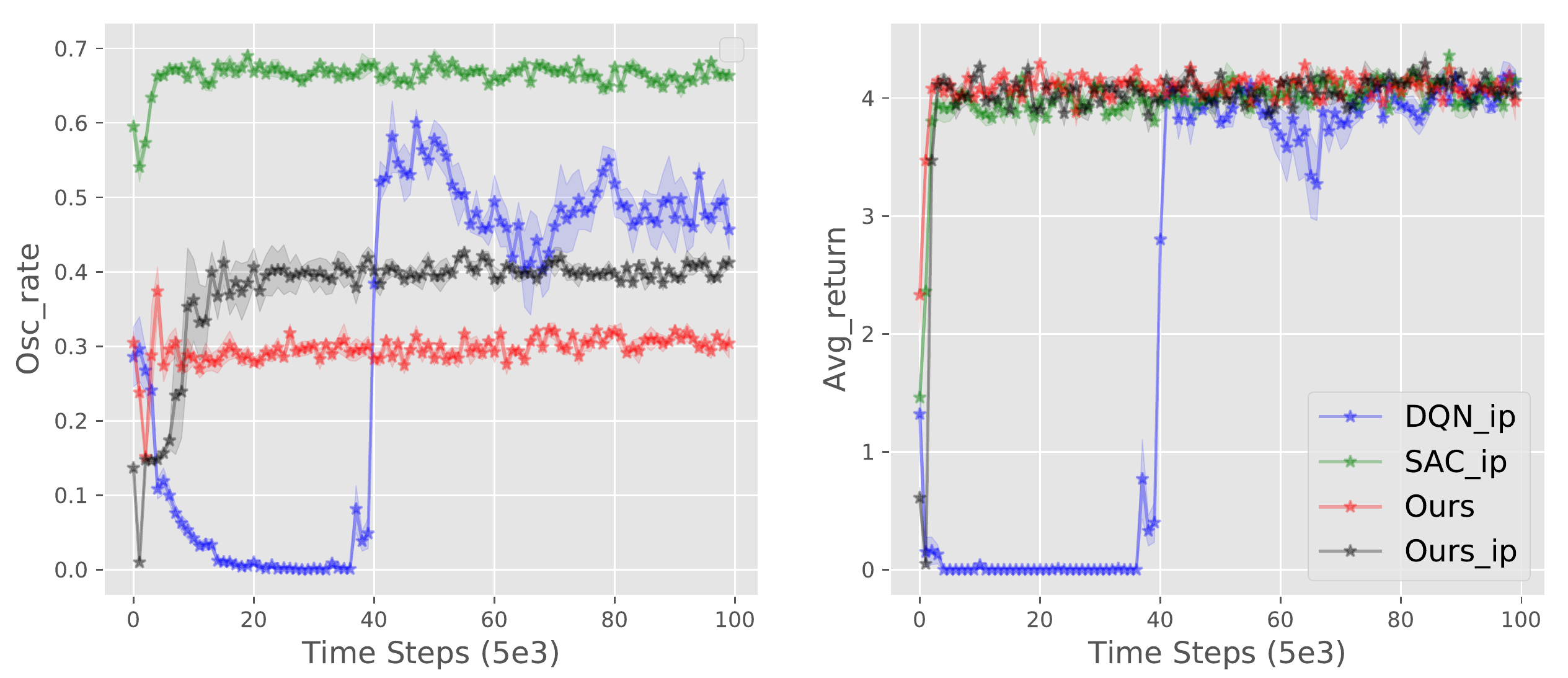}
}

\hspace{-0.3cm}
\subfigure[Two-Way]{
\label{figure:penalty_two-way}
\includegraphics[width=0.33\textwidth]{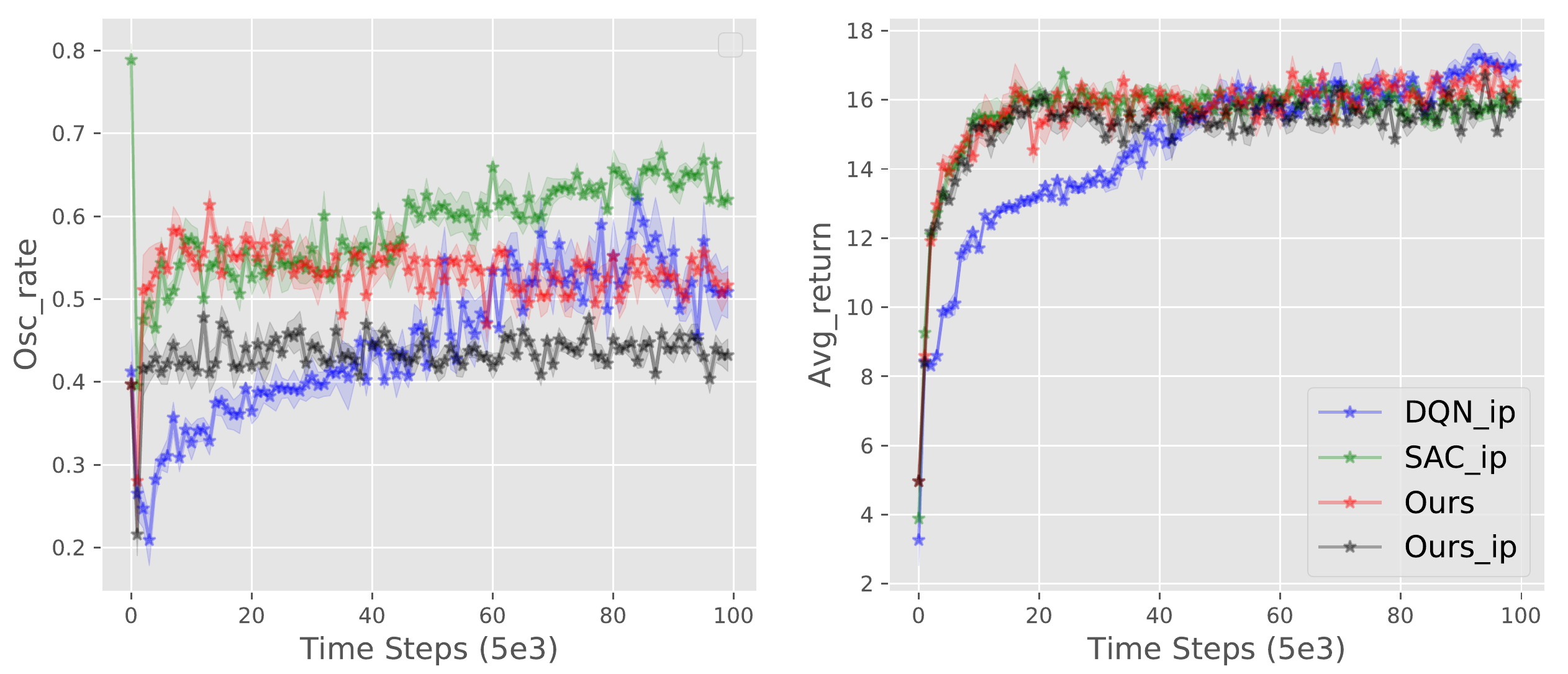}
}
\hspace{-0.3cm}
\subfigure[Complex Scenario]{
\label{figure:simple_scenario}
\includegraphics[width=0.33\textwidth]{AAAI_plot/AAAI_plot_clip/two_way_base.pdf}
}
\hspace{-0.3cm}
\subfigure[Simple Scenario]{
\label{figure:complex_scenario}
\includegraphics[width=0.33\textwidth]{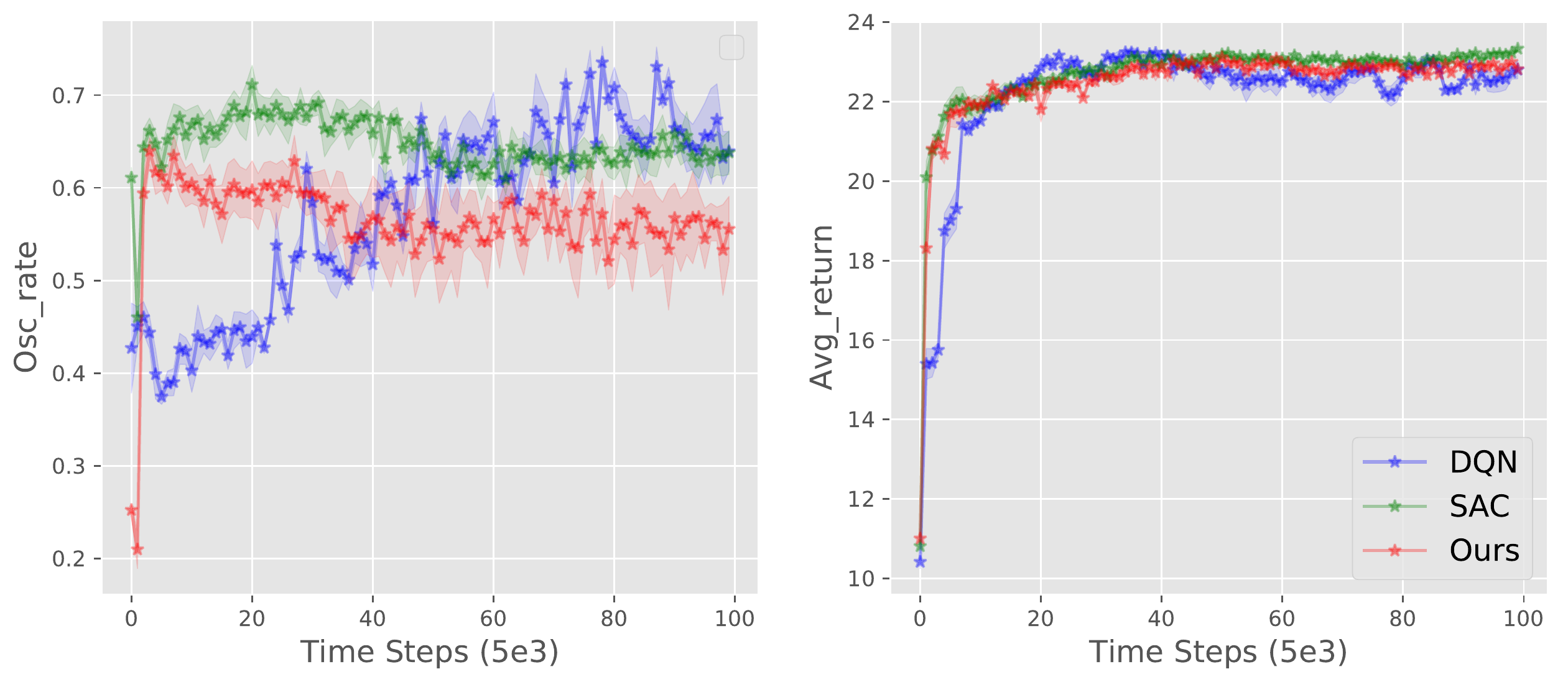}
}

\caption{Comparisons results for \emph{(a)-(b)} Action Repetition, \emph{(c)-(d)} Reward Shaping, and \emph{(e)-(f)} Complex v.s. Simple scenarios.}
\end{figure*}

\section{Experiments}
This section presents the experimental results of our approach.
We first provide the setups and the benchmark approaches in Section \ref{section:setup}, 
and then followed by evaluation results and an analysis study in Section \ref{section:evaluation_results} and Section \ref{section:analysis}.
Finally, we analyze the learned regularization of smoothness for further insights of PIC frameworks in Section \ref{section:visual}.

\subsection{Setups} 
\label{section:setup}
%\subsubsection*{Environments and Benchmark Approaches}
\textbf{Environments and Benchmark Approaches.}
We use the \emph{Highway} simulator\footnote{Highway environments are originally provided at \url{https://github.com/eleurent/highway-env}.}
which includes a collection of autonomous driving scenarios,
as well as several Atari games in OpenAI-Gym in our experiments.
Highway simulator have been used in previous works \cite{LeurentM19Practical,Luerent@Social,LiW0B19MultiView,CarraraLLUMP19Budgeted} and four typical tasks, i.e., \emph{Lane Change}, \emph{Merge}, \emph{Intersection} and \emph{Two-Way} are picked to conduct experiments. 
The state of these tasks are mainly about locomotion and kinematics of vehicles.
The action space are discrete, consisting of vehicle controls, e.g., left/right lane change, accelerate/decelerate.
%These tasks have different state spaces, action spaces, rewards degisn, episode durations and randomized system parameters and . 
Detailed configuration of the tasks are provided in  Supplementary Material C. 
%\ref{section:highway_config}.
For OpenAI Atari, we use \emph{MsPacman-v4}, \emph{SpaceInvaders-v4},  \emph{Qbert-v4} and \emph{JamesBond-v4} with pixel-input states.
We compare against benchmark approaches algorithms, including DQN \cite{Mnih2015DQN}, discrete SAC \cite{Christodoulou19SACdiscrete} and their variants with dynamic action repetition and reward shaping tricks (see Section \ref{section:evaluation_results}).
All experimental details are provided Supplementary material C.1.
%Two other settings are also considered as comparison:
%where reward is shaped with action inconsistent penalty and where action space is augmented by action repetition, 
%since these tricks are naive treatments to address the action oscillation issue.  

\textbf{Training and Evaluation.}
We train five different instances of each algorithm with different random seeds, with each performing $20$ evaluation rollouts with some other seed every $5000$ environment steps. 
For each task we report undiscounted return and the action oscillation ratio (Equation (\ref{eqation:action-oscillation-ratio})) which is calculated from the evaluation trajectories.
Concretely,  for each episode $i$, we record the count of action switch within consecutive steps, denoted by $c_i$, and the total steps of the episode, denoted by $n_i$, then oscillation ratio is computed as $(\sum_{i=1}^{20}c_i/n_i )/20$.
The solid curves corresponds to the mean and the shaded region to half a standard deviation over the five trials.

\subsection{Results}
\label{section:evaluation_results}
\textbf{Evaluations.} 
We first compare NSAC (ours) with DQN and SAC across all 4 Highway tasks and 4 Atari games.
The results are shown in Figure \ref{figure:base_results}.
We see that our approach achieves substantial reduction with respect to the action oscillation rate than benchmark approaches, especially when compared with SAC,
while retaining comparable performance across all tasks.
We credit the results to the smoothness property (Theorem \ref{thm:smoothness}) of mixed policy and the effectiveness of NPI (Theorem \ref{thm:NPI}).

%\begin{figure}\label{base_results}
%	\centering
%	\includegraphics[width=14cm]{full_results_4.pdf} 
%	\caption{Training curves of algorithms on four \emph{Highway} tasks (a)-(d) and two Atari games (e)-(f) w.r.t. action oscillation ratio (left) and returns (right). 
%	Our approach (red) consistently achieves lower action oscillation rate %while retaining comparable performance across all tasks.
%	}
%\end{figure}

\textbf{Comparison with Action Repetition.} 
Further, we absorb the core idea of action repetition works \cite{DurugkarRDM16Macro,Lakshminarayanan16bFrame,SharmaLR17Repeat} into another two variants of DQN and SAC, that learn both actions and action repetitions from extended action space. Concretely, we set the repetition set as $\mathit{Re}=\{1,2,4,8\}$ , which means that the action is repeated for $1,2,4,8$ times, then the augmented action space $\mathit{A}^{\prime}$ is the Cartesian product $\mathit{A} \times \mathit{Re}$. \emph{DQN-repeat} baseline and \emph{SAC-repeat} baseline mean that DQN and SAC are trained on the augmented action space $\mathit{A}^{\prime}$, respectively.
We present representative comparison results in \emph{Intersection} and \emph{Two-Way} as shown in Figure \ref{figure:repeat_intersection} -\ref{figure:repeat_two-way}.
The results show that the action repetition approaches can achieve certain reduction in action oscillation yet sacrificing performance when comparing with our approach within the same environment steps.
This is because action repetition hampers sample efficiency due to temporal abstraction as we discuss before.

\textbf{Comparison with Reward Shaping.} 
Moreover, we also consider the setting where the reward is shaped with action inconsistent penalty, to some degree, this can be viewed to be equivalent to inject a regularizer based on $negative$ action oscillation ration defined in Equation \ref{eqation:action-oscillation-ratio}. \emph{DQN-ip}, \emph{SAC-ip} and \emph{Ours-ip} are DQN, SAC, NSAC algorithms trained on the environments with action inconsistency penalty -0.05 within consecutive steps yet evaluated without it. 
The results in Figure \ref{figure:penalty_intersection}-\ref{figure:penalty_two-way} show such reward shaping is effective in reducing oscillation in \emph{Two-Way} for all DQN, SAC and our approach, while cause a counterproductive result in \emph{Intersection}. 
We conjecture it is because action inconsistent penalty violates the original reward structure (sparse reward in \emph{Intersection}) then the learned policies tend to fall into the unexpected local minimum, revealing the poor scalability of such reward shaping treatment.
Additionally, reward shaping can be exhaustive and even impossible in complex problems. 
% Complete comparison results on more tasks are provided in Supplementary Material D.
%\ref{section:complete_results}.

%On the tasks with sparse goal-oriented reward siginals such as ``Merge", NSAC shows dominent advantages since
%reward shaping method  is  counter-productive in the oscillation rate for both our method and  baseline algorihtms. 
%We conjecture it is because action inconsistent penalty  violates the original reward structure then the learned policies tend to fall into the unexpected local minumum, revealing that reward shaping method is not a scalable way in principle. 
%On  the tasks with dense reward signals as shown in Fig. \ref{ip} (b),  Our method generates comparable action oscillation reduction when comparing with reward shaped baselines. %More than that, NSAC can also be used in combinition with reward shaping method and then reinforced effects are achieved with the extra action oscillation reduction.

% \begin{figure}\label{ip}
% \centering
% \subfigure[intersection-v0]{
% \includegraphics[width=0.4\textwidth]{plots/intersection_ip.pdf}
% }
% \hspace{-0.8cm}
% \subfigure[two-way-v0]{
% \includegraphics[width=0.4\textwidth]{plots/two_way_ip.pdf}
% }

% \caption{Reward is shaped with action inconsistent penalty. The oscillation rates do not drop on Intersection task.}
% %\caption{Cooperative games.}
% \label{figure:2}
% \end{figure}

\begin{figure*}
	\centering
	\includegraphics[width=0.9\textwidth]{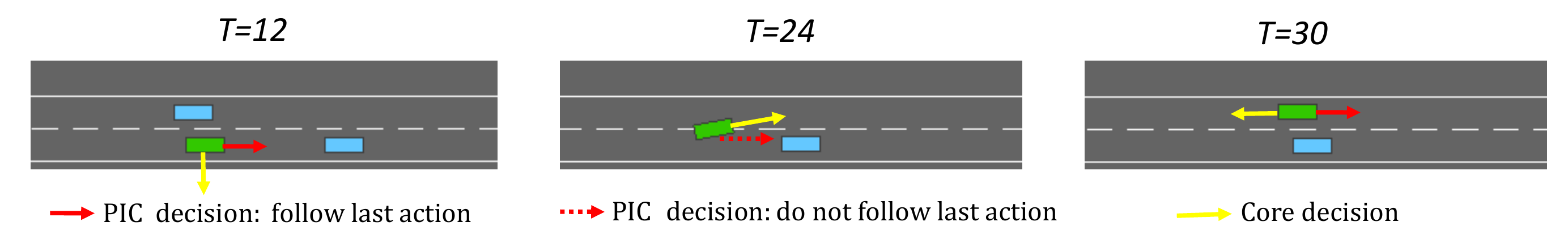} 
	\caption{Visualization of several consecutive frames in \emph{Two-Way}.
	Green denotes the vehicle controlled by the mixed policy learned with NSAC and blue denotes other vehicles to overtake.
	Red and yellow arrows illustrates the decisions of the PIC module and policy core.
	The solid arrow denote a high value or probability ($\ge 0.5$) and the dashed is for a low one.
	We see that the PIC module gives good regulation when improper oscillations happen ($t=12$ and $t=14$) while impose few regulation to follow the previous action for necessary change ($t=24$).}
	\label{figure:visual}
\end{figure*}

\subsection{Analysis Study}
\label{section:analysis}
%The results in the previous section suggest that our algorithm  NSAC can outperform conventional off policy RL methods in terms of oscillation rate almost without any performance loss. 
In this section, we conduct several analysis studies to further examine our approach from three aspects:  the performance  in complex scenario against simple scenario, the influence of an extra lower bound on the output of PIC module $\mu^{\rm pic}$ and  the effect of   temperature parameter $\alpha$ of the mixed policy $\pi$.
%we further examine which particular component are important for good performance. 
%We also examine how sensitive our algorithm is to some of the most important hyperparameters, namely the lower bound of inertia controller $p_0$ and the temparature parameter $\alpha^{\rm mix}$ of the mixed policy $\pi$.

% \begin{wrapfigure}{r}{0cm}
% \begin{minipage}[t]{.38\linewidth}
% \centering
% \hspace{-0.7cm}
% \subfigure[Complex Scenario]{
% \label{figure:hierarchical_control}
% \includegraphics[width=1.1\textwidth]{plots/ablation_complex.pdf}
% }

% \hspace{-0.7cm}
% \subfigure[Simple Scenario]{
% \label{figure:hierarhical_comparison}
% \includegraphics[width=1.1\textwidth]{plots/ablation_simple.pdf}
% }

% \end{minipage}
% \caption{
% sss
% }
% \label{figure:complex}
% \end{wrapfigure}

\textbf{Simple v.s. Complex Scenarios.}
To compare how the complexity of the environments affects the performance, 
we conduct experiments on both the complex scenarios and simple scenarios on \emph{Two-way} task, where the vehicles number in complex scenarios doubles that in simple scenarios. 
Figure \ref{figure:simple_scenario}-\ref{figure:complex_scenario} indicates that the oscillation reduction is more significant in complex cases with a large margin.
This is as expected that the policy learned is likely to be more bumpy since the solution policy space become more complex, and thus there exists a larger space for our approach to reduce the oscillation in actions.

\begin{figure}
	\centering
	\includegraphics[width=0.45\textwidth]{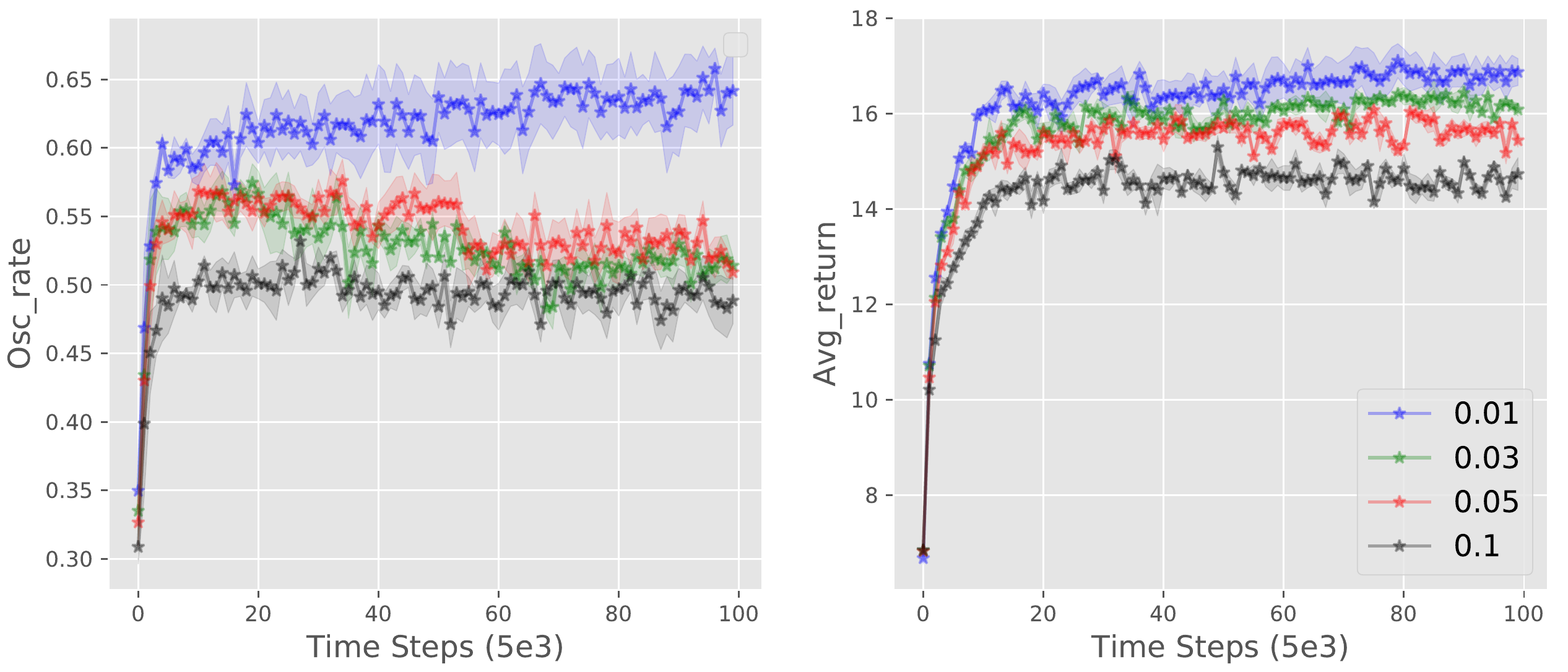} 
	\caption{Training curves of our algorithm on the complex scenario of \emph{Two-Way} with different lower bound values $\mu_0$ of the policy inertia controller $\mu^{\rm pic}$.}
	\label{figure:complex_p0}
\end{figure}

\textbf{Lower bound of Policy Inertia Controller.}
We also consider to impose an extra lower bound $\mu_0$ on the PIC module $\mu^{\rm pic}$ to further encourage smoothness of learned policies.
We find in Figure \ref{figure:complex_p0} that
a smaller $\mu_0$ performs better
regarding both oscillation rate and average return,
%since it encourages the exploration of smooth policies 
while a large $\mu_0$ induces too much regulation which causes substantial oscillation reduction but performance degradation as well.

\begin{figure}
	\centering
	\includegraphics[width=0.45\textwidth]{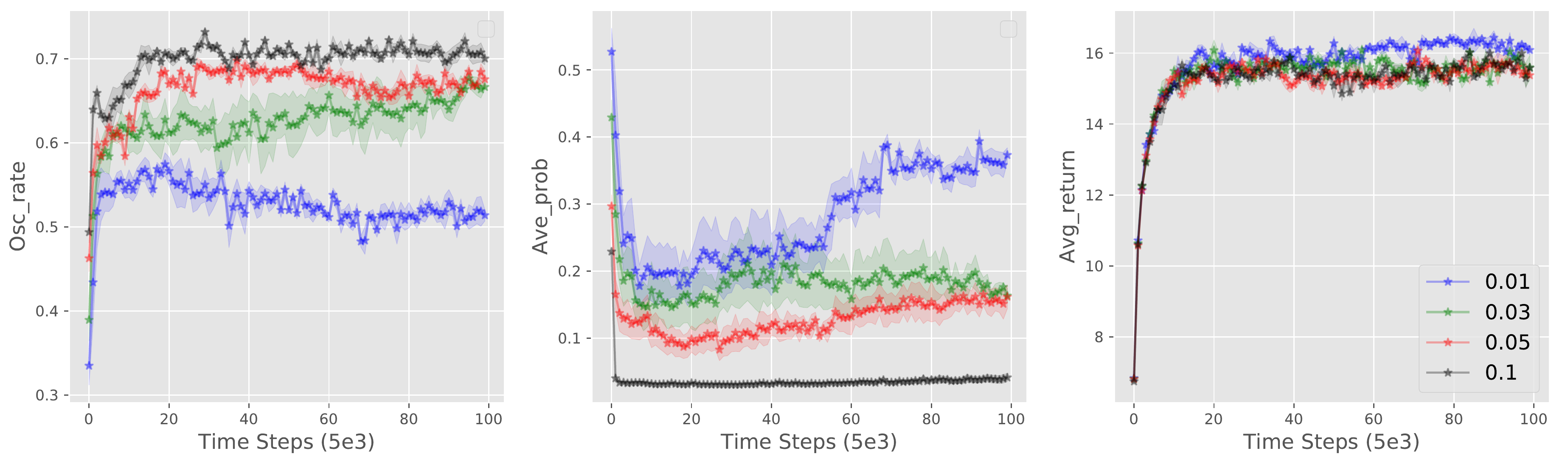} 
	\caption{Training curves of our algorithm on \emph{Two-Way} with different temperature parameter $\alpha$ of the mixed policy $\pi$.
	The results are with respect to action oscillation ratio (left), average inertia controller's outputs (middle) and returns (right).
	A smaller $\alpha$ induces better regularization (middle), lower action oscillation rate (left) as well as a higher performance (right).}
	\label{figure:almi}
\end{figure}

\textbf{Temperature Parameter of the Mixed Policy.}
We additionally examine the influence of the temperature parameter $\alpha$ of the mixed policy $\pi$.
The results in Figure \ref{figure:almi}
show that a smaller $\alpha$ induces better regularization of $\mu^{\rm pic}$ (which is computed as the average $\mu^{\rm pic}$ in an episode),
lower action oscillation rate as well as a higher performance.
A relatively large $\alpha$ hampers the regularization of $\mu^{\rm pic}$
since it encourages the stochasticity of the mixed policy.

% Complete results and other analysis study can be found in Supplementary Material D.
%\ref{section:complete_results}.

\subsection{A Close Look at Learned PIC Regularization}
\label{section:visual}
To better understand how the regulation on policy core $\pi^{\rm core}$ is given by a learned PIC module $\mu^{\rm pic}$, we visualize the execution of the vehicle (green) in \emph{Two-Way} controlled by the mixed policy learned with NSAC, as in Figure \ref{figure:visual}.
We find an oscillation occurs at timestep $t=12$ when $\pi^{\rm core}$ tend to pick the right lane change action, and $\mu^{\rm pic}$ outputs a high value to keep the vehicle forward.
At timestep $t=24$, when $\pi^{\rm core}$ chooses the left lane change action, $\mu^{\rm pic}$ outputs a low value that does not regulate $\pi^{\rm core}$ to keep forward any more.
At timestep $t=30$, another improper oscillation happens when the decelerate action is encouraged by $\pi^{\rm core}$, $\mu^{\rm pic}$ regulates it to keep accelerating instead.
This shows 
the PIC module has learned a good strategy to remedy improper oscillations ($t=12$ and $t=14$) by strongly regulate policy core to follow the previous action, while to impose few regulation when necessary change ($t=24$) is offered by policy core.
%Placeholder

\section{Conclusion}
In this paper, we propose a generic framework Policy Inertia Controller (PIC) to address the action oscillation issue of DRL algorithms through directly regulating the policy distribution.
Moreover, we propose Nested Policy Iteration to train the PIC-augmented policies with monotonically non-decreasing updates in a general way.
%We also introduce PIC-DSAC as a representative example that ensures monotonically non-decreasing updates.
Our empirical results in a range of autonomous driving tasks and several Atari games show that our derived Nested Soft Actor-Critic algorithm achieves substantial action oscillation reduction without sacrificing policy performance at the same time, which is of much significance to real-world scenarios.
The future work is to apply and develop our approaches in practical applications like real-world autonomous driving cars, and to investigate the extension for continuous policies.

\section*{Acknowledgments}
The work is supported by the National Natural Science Foundation of China (Grant Nos.: 61702362, U1836214, U1813204), Special Program of Artificial Intelligence and Special Program of Artificial Intelligence of Tianjin Municipal Science and Technology Commission (No.: 56917ZXRGGX00150), Tianjin Natural Science Fund (No.: 19JCYBJC16300), Research on Data Platform Technology Based on Automotive Electronic Identification System.

\bibliography{aaai21_PIC}

\clearpage

\onecolumn
\appendix

\section{Complete proofs}

\subsection{Proof of Theorem \ref{thm:smoothness}}
\label{proof:smoothness}
% \begin{lem}
% Assume that there exist two real sequences $\mathbf{\alpha} =\{\alpha_{0},\alpha_{1},\cdots,\alpha_{n}\}$ with $0 \leq\alpha_i \leq 1, \forall i$, and $\mathbf{\beta} =\{\beta_{0},\beta_{1},\cdots,\beta_{n}\}$ with $0 \leq\beta_i \leq 1, \forall i$. We also assume a random sequence  $\{o_0,\cdots, o_n\} $ where  $o_i$  obeys a Bernoulli distribution with the parameter $p_i$. Define two sequences $\{a_k, k=0,\cdots,n\}$ and  $\{c_k,k=0,\cdots,n\}$  with $a_k: = \sum_{\{o_i\}:\sum_{i}o_i=k}\prod_{i=0}^{n}\alpha_{i}$ and  $c_k:=\sum_{\{o_i\}:\sum_{i}o_i=k}\prod_{i=0}^{n}(\alpha_{i}+\beta_i)^{o_i}\big((1-\beta_{i})\alpha_i\big)^{1-o_i}$  then we have
% $\{c_k - a_k, k=0,\cdots n\}$ is monotonically increasing.

% \end{lem}

\textbf{Proof:} We define an auxiliary reward function  $r_{o}(s_t,a_{t-1},a_t) = r_{o}(a_t,a_{t-1}) =  \mathbb{I}_{\{a_{t-1}\}}(a_t)$ and a default action $a_{-1}$ as the previous action of $a_0$ specially to make the formulation work. We also define $V_{o}^{\pi}(s)$ as the value function of the state $s$ under the reward function $r_{o}(s_t,a_{t-1},a_t)$ for given $\pi$.
Therefore, to prove $\xi(\pi) \le \xi(\pi^{\rm core})$ is equivalent to prove $\mathbb{E}_{s_0 \sim \rho_0} \left[ \frac{1}{T} V_{o}^{\pi}(s_0) \right] \geq \mathbb{E}_{s_0 \sim \rho_0} \left[ \frac{1}{T} V_{o}^{\pi^{\rm core}}(s_0) \right]$, 
thus to prove $V_{o}^{\pi}(s_0) \geq V_{o}^{\pi^{\rm core}}(s_0)$ for all possible initial states $s_0$ given that $\gamma=1$. 
	
% For $s_0$, we expand $V_{o}^{\pi^{\rm core}}$ as follows:
% 	\begin{eqnarray}
% 	V_{o}^{\pi^{\rm core}}(s_0) 
% 	&=& \mathbb{E}_{a_0\sim \pi^{\rm core}(\cdot|s_0)}\big[r_{o}(s_0,a_0)+  \mathbb{E}_{s_1\sim p(\cdot|s_0,a_0)}V_{o}^{\pi^{\rm core}}(s_1)    \big]  \nonumber \\
% 	&=& \mathbb{E}_{a_0\sim \pi^{\rm core}(\cdot|s_0)}r_{o}(s_0,a_0) + \mathbb{E}_{a_0\sim \pi^{\rm core}(\cdot|s_0)}\mathbb{E}_{s_1\sim p(\cdot|s_0,a_0)}V_{o}^{\pi^{\rm core}}(s_1). 
% 	\end{eqnarray}	
    %By the definition of $r_{o}$, for any $\mu(s_0,a_{-1}) > 0$ we have
    %\begin{eqnarray}
    %&~& \mathbb{E}_{a_0\sim \pi^{\rm core}(\cdot|s_0)}\big[r_{o}(s_0,a_0)\big] \nonumber \\
    %& <&(1-\mu(s_0,a_{-1})) \mathbb{E}_{a_0\sim \pi^{\rm core}(\cdot|s_0)}\big[r_{o}(s_0,a_0)\big] + \mu(s_0,a_{-1})r_{o}(a_0,a_{-1}) \nonumber \\
    %&=& \mathbb{E}_{a_0\sim \pi(\cdot|s_0)}\big[r_{o}(s_0,a_0)\big].
     %\end{eqnarray}
    For $s_0$ and arbitrary $\mu(s_0,a_{-1}) \geq 0 $ with its corresponding $\pi$, we have
    \begin{eqnarray}
    &~& \mathbb{E}_{a_0\sim \pi(\cdot|s_0)} Q_o^{\pi^{\rm core}}(s_0, a_{-1}, a_0) \nonumber \\
    &=& \mathbb{E}_{a_0\sim \pi(\cdot|s_0)} \left[r_{o}(s_0,a_{-1},a_0) + \mathbb{E}_{s_1\sim p(\cdot|s_0,a_0)}V_{o}^{\pi^{\rm core}}(s_1) \right] \nonumber \\
    &=&\mathbb{E}_{a_0\sim \pi(\cdot|s_0)}r_{o}(s_0,a_{-1},a_0) + \mathbb{E}_{a_0\sim \pi(\cdot|s_0)}\mathbb{E}_{s_1\sim p(\cdot|s_0,a_0)}V_{o}^{\pi^{\rm core}}(s_1) \nonumber \\
    &=&(1-\mu(s_0,a_{-1}))\underbrace{ \mathbb{E}_{a_0\sim \pi^{\rm core}(\cdot|s_0)}\big[r_{o}(s_0,a_{-1},a_0)\big]}_{\text{A}} + \mu(s_0,a_{-1})r_{o}(a_0,a_{-1}) \nonumber \\
    &+&
    (1-\mu(s_0,a_{-1}))\underbrace{\mathbb{E}_{a_0\sim \pi^{\rm core}(\cdot|s_0)}\mathbb{E}_{s_1\sim p(\cdot|s_0,a_0)}V_{o}^{\pi^{\rm core}}(s_1)}_{\text{B}} +\mu(s_0,a_{-1})\underbrace{\mathbb{E}_{s_1\sim p(\cdot|s_0,a_{-1})}V_{o}^{\pi^{\rm core}}(s_1)}_{\text{C}} \nonumber \\
    & =& A+B+\mu(s_0,a_{-1})(1-A+C-B) = V_{o}^{\pi^{\rm core}}(s_0)  + \mu(s_0,a_{-1})(1-A+C-B) \label{theorem:2}
   % &=& \mu(s_0,a_{-1}) + (1-\mu(s_0,a_{-1}))A +(1-\mu(s_0,a_{-1}))B+ \mu(s_0,a_{-1})C \geq 0, \label{theorem:2}
    \end{eqnarray}
    where the last equality comes from the definition of $r_{o}$ and $1-A\geq 0$ always holds. 
    
    % Similarly, we can expand the standard expand $V^{\pi^{\rm core}}$ for the original reward function $r(s_t,a_t)$ as follows:
    % 	\begin{eqnarray}
    % V^{\pi^{\rm core}}(s_0) 
    % &=& \mathbb{E}_{a_0\sim \pi^{\rm core}(\cdot|s_0)}\big[r(s_0,a_0)+  \mathbb{E}_{s_1\sim p(\cdot|s_0,a_0)}V^{\pi^{\rm core}}(s_1)    \big]  \nonumber \\
    % &=& \mathbb{E}_{a_0\sim \pi^{\rm core}(\cdot|s_0)}r(s_0,a_0) + \mathbb{E}_{a_0\sim \pi^{\rm core}(\cdot|s_0)}\mathbb{E}_{s_1\sim p(\cdot|s_0,a_0)}V^{\pi^{\rm core}}(s_1). 
    % \end{eqnarray}
   We also have the similar relation on the original reward $r(s_t,a_t)$ as follows:
    \begin{eqnarray}
    &~& \mathbb{E}_{a_0\sim \pi(\cdot|s_0)} Q^{\pi^{\rm core}}(s_0, a_0) \nonumber \\
    &=& \mathbb{E}_{a_0\sim \pi(\cdot|s_0)} \left[r(s_0,a_0) + \mathbb{E}_{s_1\sim p(\cdot|s_0,a_0)}V^{\pi^{\rm core}}(s_1) \right] \nonumber \\
    &=&\mathbb{E}_{a_0\sim \pi(\cdot|s_0)} r(s_0,a_0) + \mathbb{E}_{a_0\sim \pi(\cdot|s_0)}\mathbb{E}_{s_1\sim p(\cdot|s_0,a_0)}V^{\pi^{\rm core}}(s_1) \nonumber \\
    &=&(1-\mu(s_0,a_{-1}))\underbrace{ \mathbb{E}_{a_0\sim \pi^{\rm core}(\cdot|s_0)}\big[r(s_0,a_0)\big]}_{\text{A}^{\prime}} + \mu(s_0,a_{-1})r(s_0,a_{-1}) \nonumber \\
    &+&
    (1-\mu(s_0,a_{-1}))\underbrace{\mathbb{E}_{a_0\sim \pi^{\rm core}(\cdot|s_0)}\mathbb{E}_{s_1\sim p(\cdot|s_0,a_0)}V^{\pi^{\rm core}}(s_1)}_{\text{B}^{\prime}} +\mu(s_0,a_{-1})\underbrace{\mathbb{E}_{s_1\sim p(\cdot|s_0,a_{-1})}V^{\pi^{\rm core}}(s_1)}_{\text{C}^{\prime}} \nonumber \\
    & =& A'+B'+\mu(s_0,a_{-1})(r(s_0,a_{-1})-A'+C'-B') \nonumber \\
   & =& V^{\pi^{\rm core}}(s_0) + \mu(s_0,a_{-1})(r(s_0,a_{-1})-A'+C'-B')
   \label{theorem:3}
    % &=& \mu(s_0,a_{-1}) + (1-\mu(s_0,a_{-1}))A +(1-\mu(s_0,a_{-1}))B+ \mu(s_0,a_{-1})C \geq 0, \label{theorem:2}
    \end{eqnarray}
Hence, we can adopt the following rule to pick $\mu(s_0,a_{-1})$: 
\begin{eqnarray}
\mu(s_0,a_{-1})\left\{
\begin{aligned}
& >  0,~  \text{if}~& (1-A)+(C-B)>0 ~\text{and}~ r(a_0,a_{-1})-A'+C'-B'>0 \\
& =  0,~  \text{else}.&
\end{aligned}
\right.
\end{eqnarray}
By this way, we can ensure that (\ref{theorem:2}) is not less than $V_{o}^{\pi^{\rm core}}(s_0)$ and (\ref{theorem:3}) is not less than $V^{\pi^{\rm core}}(s_0)$ as well.
And this implies  that
    \begin{eqnarray}
    V_o^{\pi^{\rm core}}(s_0) \le \mathbb{E}_{a_0\sim \pi(\cdot|s_0)} Q_o^{\pi^{\rm core}}(s_0, a_{-1}, a_0)
    %&~&\mathbb{E}_{a_0\sim \pi^{\rm core}(\cdot|s_0)}r_{o}(s_0,a_0) + \mathbb{E}_{a_0\sim \pi^{\rm core}(\cdot|s_0)}\mathbb{E}_{s_1\sim p(\cdot|s_0,a_0)}V_{o}^{\pi^{\rm core}}(s_1) \nonumber \\
    %&\leq& \mathbb{E}_{a_0\sim \pi(\cdot|s_0)}r_{o}(s_0,a_0) + \mathbb{E}_{a_0\sim \pi(\cdot|s_0)}\mathbb{E}_{s_1\sim p(\cdot|s_0,a_0)}V_{o}^{\pi^{\rm core}}(s_1), 
    \label{theorem:1}
    \end{eqnarray}
and 
    \begin{eqnarray}
    V^{\pi^{\rm core}}(s_0) \le \mathbb{E}_{a_0\sim \pi(\cdot|s_0)} Q^{\pi^{\rm core}}(s_0, a_0). 
    %&~&\mathbb{E}_{a_0\sim \pi^{\rm core}(\cdot|s_0)}r(s_0,a_0) + \mathbb{E}_{a_0\sim \pi^{\rm core}(\cdot|s_0)}\mathbb{E}_{s_1\sim p(\cdot|s_0,a_0)}V^{\pi^{\rm core}}(s_1) \nonumber \\
    %&\leq& \mathbb{E}_{a_0\sim \pi(\cdot|s_0)}r(s_0,a_0) + \mathbb{E}_{a_0\sim \pi(\cdot|s_0)}\mathbb{E}_{s_1\sim p(\cdot|s_0,a_0)}V^{\pi^{\rm core}}(s_1). 
    \label{theorem:4}
  \end{eqnarray}  
	Repeatedly expand $V_{o}^{\pi^{\rm core}}$  on the RHS by applying the inequality (\ref{theorem:1}) and follow the rule of choosing $\mu(s_t,a_{t-1})$ at each state,
	we have
	\begin{eqnarray}
	 V_{o}^{\pi^{\rm core}}(s_0)
	&\le& \mathbb{E}_{a_0\sim \pi(\cdot|s_0)} Q_o^{\pi^{\rm core}}(s_0, a_{-1},a_0) \nonumber \\
	&=& \mathbb{E}_{a_0\sim \pi(\cdot|s_0)} \left[r_{o}(s_0,a_{-1},a_0) + \mathbb{E}_{s_1\sim p(\cdot|s_0,a_0)}V_{o}^{\pi^{\rm core}}(s_1) \right] \nonumber \\
	&\le& \mathbb{E}_{a_0\sim \pi(\cdot|s_0)}
	\left[r_{o}(s_0,a_{-1},a_0)
	+ \mathbb{E}_{s_1\sim p(\cdot|s_0,a_{-1},a_0)}\bigg( \mathbb{E}_{a_1\sim \pi(\cdot|s_1)}r(s_1,a_0,a_1) + 
	\mathbb{E}_{a_1\sim \pi(\cdot|s_1)}\mathbb{E}_{s_2 \sim p^{\pi}(\cdot|s_1,a_1)}V_{o}^{\pi^{\rm core}}(s_2)   \bigg)
	\right]
	\nonumber \\
	&\leq& \vdots \nonumber \\
	&\leq& V_{o}^{\pi}(s_0). \label{theorem:5}
	\end{eqnarray}
	Similarly, we can  repeatedly expand $V^{\pi^{\rm core}}$  on the RHS by applying the inequality (\ref{theorem:4}) and obtain
	\begin{eqnarray}
V^{\pi^{\rm core}}(s_0) \leq V^{\pi}(s_0).\label{theorem:6}
\end{eqnarray}	
	%where we have repeatedly expand $V^{\pi}$ on the RHS by applying the inequality (\ref{proof_theorem_2}). 
(\ref{theorem:5}) and (\ref{theorem:6}) yield  Theorem \ref{thm:smoothness}.

\subsection{Proof of Lemma \ref{lemma:intermidiate_policy_improvement}}
\label{proof:pic-improvement}
Two lemmas are presented firstly.
\begin{lem}\cite{schulman2015trust}\label{trpo_lemma1}
For arbitrary $\theta$ and $\theta'$, the following equality holds:
\begin{eqnarray}
J(\theta')-J(\theta) = \mathbb{E}_{\tau \sim p_{\theta'}(\tau)}\bigg[  \sum_{t}\gamma^t A^{\pi_{\theta}}(s_t,a_t) \bigg].
\end{eqnarray}
\end{lem}
The following lemma can be derived from \cite{schulman2015trust}:
\begin{lem}
    \label{trpo}
	Assume $\pi(s_t)$ is an arbitrary  distribution, then 
	if $|\pi'(a_t|s_t) - \pi(a_t|s_t)|< \epsilon $  for all $s_t,a_t$, we have
	$\pi'(s_t)$ is close enough to $\pi(s_t)$ , and the following results hold:
	\begin{eqnarray}
	|p_{\pi'}(s_t) -p_{\pi}(s_t)| \leq 2 \epsilon t, \label{trpo_eqa_1}
	\end{eqnarray}
	\begin{eqnarray}
	\mathbb{E}_{p_{\pi'}(s_t)}[f(s_t)] \geq \mathbb{E}_{p_{\pi}(s_t)}[f(s_t)] -2\epsilon t \max_{t}|f(s_t)|,\label{trpo_eqa_2}
	\end{eqnarray}
	where $p_{\pi}$ denotes the state marginal of the trajectory distribution induced by policy $\pi$ (in a specific MDP).
\end{lem}

\textbf{Proof of Lemma \ref{lemma:intermidiate_policy_improvement}:} 
Firstly, we   rewrite $Q_{\rm mid}(s_t,a_{t-1},a_t)-Q_{\rm old}(s_t,a_{t-1},a_t)$ into three parts as follows:
\begin{eqnarray}
&~& Q_{\rm mid}(s_t,a_{t-1},a_t)-Q_{\rm old}(s_t,a_{t-1},a_t) \nonumber \\
&=& \bigg(Q_{\rm mid}(s_t,a_{t-1},a_t) - Q^{\rm core}_{\rm new}(s_t,a_{t-1},a_t)\bigg)
+  \bigg(Q^{\rm core}_{\rm new}(s_t,a_t) -  Q^{\rm core}_{\rm old}(s_t,a_t)\bigg) \nonumber \\
&~~~~~~~~& +  \bigg(Q^{\rm core}_{\rm old}(s_t,a_{t-1},a_t) - Q_{\rm old}(s_t,a_{t-1},a_t)\bigg), \label{three_parts}
\end{eqnarray}
where we have used the fact that 
$Q^{\rm core}_{\rm new}(s_t,a_{t-1},a_t) = Q^{\rm core}_{\rm new}(s_t,a_t) $ and $Q^{\rm core}_{\rm old}(s_t,a_{t-1},a_t) = Q^{\rm core}_{\rm old}(s_t,a_t) $.

We expand the first term in (\ref{three_parts}) as follows:
\begin{eqnarray}
&~& Q_{\rm mid}(s_t,a_{t-1},a_t) - Q^{\rm core}_{\rm new}(s_t,a_{t-1},a_t) \nonumber \\
&=& \mathbb{E}_{s_{t+1}\sim p(s_{t+1}|s_t,a_t)}[V^{\pi_{\rm mid}}(s_{t+1},a_{t})] - \mathbb{E}_{s_{t+1}\sim p(s_{t+1}|s_t,a_t)}[V^{\pi_{\rm new}^{\rm core}}(s_{t+1},a_{t})]  \nonumber  \\
&=& \mathbb{E}_{\tau\sim \pi_{\rm mid}}[V^{\pi_{\rm mid}}(s_t,a_{t-1})] - \mathbb{E}_{{\tau\sim \pi^{\rm core}_{\rm new}}}[V^{\pi^{\rm core}_{\rm new}}(s_t,a_{t-1})] \nonumber    \\
&=& \mathbb{E}_{\tau\sim \pi_{\rm mid}}[\sum_{t}\gamma^{t}A^{\pi^{\rm core}_{\rm new}}(s_t,a_{t-1},a_t)], \label{first_part}
\end{eqnarray}
where the last equality comes from Lemma \ref{trpo_lemma1}.
Since the difference of action probability between $\pi_{\rm mid}(\cdot|s_t,a_{t-1})$
and $\pi^{\rm core}_{\rm new}(\cdot|s_t,a_{t-1})$  is bounded by $\mu_{\rm old}^{\rm pic}(s_t,a_{t-1})$, and 
$\mu_{\rm old}^{\rm pic}(s_t,a_{t-1})$ is also uniformly upper bounded by $\epsilon := \frac{\min_{(s_t,a_t)} (Q^{\rm core}_{\rm new}(s_t,a_t) - Q^{\rm core}_{\rm old}(s_t,a_t))}{N  \cdot C_0(\sum_{t}^{T}t\gamma^{t})} $ from the conditions in Lemma \ref{lemma:intermidiate_policy_improvement}, then directly applying (\ref{trpo_eqa_2}) in Lemma \ref{trpo} we obtain
\begin{eqnarray}
&~& \mathbb{E}_{\tau\sim \pi_{\rm mid}}[\sum_{t}\gamma^{t}A^{\pi_{\rm new}^{\rm core}}(s_t,a_t)] 
\geq \mathbb{E}_{\tau\sim \pi_{\rm new}^{\rm core}}[\sum_{t}\gamma^{t}A^{\pi_{\rm new}^{\rm core}}(s_t,a_t)]
-\sum_{t}2\epsilon C_0t\gamma^{t}  \nonumber \\
&=&  - \frac{2}{N} \min_{(s_t,a_t)} (Q^{\rm core}_{\rm new}(s_t,a_t) - Q^{\rm core}_{\rm old}(s_t,a_t)),    \label{second_part}
\end{eqnarray}
where $C_0$ is the upper bound of both $A^{\pi_{\rm new}^{\rm core}}$ and $A^{\pi_{\rm old}}$, 
the equality holds because of the fact that $\forall \pi, \mathbb{E}_{\tau\sim \pi}[\sum_{t}\gamma^{t}A^{\pi}(s_t,a_t)] =0$.

For the third term we can similarly derive that
\begin{eqnarray}
&~& Q^{\rm core}_{\rm old}(s_t,a_{t-1},a_t) - Q_{\rm old}(s_t,a_{t-1},a_t)  = \mathbb{E}_{\tau\sim \pi^{\rm core}_{\rm old}}[\sum_{t}\gamma^{t}A^{\pi_{\rm old}}(s_t,a_t)] \nonumber \\
&\geq & -\frac{2}{N} \min_{(s_t,a_t)} (Q^{\rm core}_{\rm new}(s_t,a_t) - Q^{\rm core}_{\rm old}(s_t,a_t)).
\label{third_part}
\end{eqnarray}

Combining (\ref{first_part}) and (\ref{third_part}) with (\ref{three_parts}), we obtain
\begin{eqnarray}
&~& Q_{\rm mid}(s_t,a_{t-1},a_t)-Q_{\rm old}(s_t,a_{t-1},a_t) \nonumber \\
&\geq& (1-\frac{4}{N})\min_{(s_t,a_t)} (Q^{\rm core}_{\rm new}(s_t,a_t) - Q^{\rm core}_{\rm old}(s_t,a_t)),
\end{eqnarray}
which yields our proof.

\section{Overall Algorithm of Nested Soft Actor-Critic (NSAC)}
\label{section:overall_alg}
We propose Nested Soft Actor-Critic (NSAC) to train the PIC-augmented policy (i.e., the mixed policy $\pi$), consisting of the PIC module $\mu^{\rm pic}$ and the policy core $\pi^{\rm core}$ simultaneously.
NSAC is a practical implementation of Nested Policy Iteration (NPI) we proposed in Section \ref{section:NPI}.
For both the inner policy iteration and the outer policy iteration, we use SAC algorithm.

For the inner policy iteration (i.e., to train the policy core $\pi^{\rm core}(\cdot|s_t)$), 
%we parameterize the policy core $\pi^{\rm core}_{\phi}(\cdot|s_t)$ with parameter $\phi$ and its $Q$-function $Q^{\rm core}_{\theta}(s_t,a_t)$ with parameter $\theta$ respectively.
we parameterize the policy core and its $Q$-function as $\pi^{\rm core}_{\phi}(\cdot|s_t)$ and $Q^{\rm core}_{\theta}(s_t,a_t)$ with parameters $\phi$ and $\theta$ respectively.
We also set the temperature parameter for training $\pi^{\rm core}(\cdot|s_t)$ as a hyperparameter $\alpha^{\rm core}$.
Then $Q^{\rm core}_{\theta}(s_t,a_t)$ is trained to minimize the soft Bellman residual according to Equation (\ref{soft_q_definition}):
\begin{eqnarray}
\mathcal{L}^{\rm core}(\theta)=E_{(s_t,a_t)\sim D}\bigg[\frac{1}{2} \big( Q^{\rm core}_{\theta}(s_t,a_t)- (r(s_t,a_t)+\gamma E_{s_{t+1}\sim P}[V^{\rm core}_{\bar{\theta}}(s_{t+1})])   \big)^2         \bigg], 
\label{J(theta)}
\end{eqnarray}  
where $D$ is a replay buffer of past experiences collected using the mixed policy $\pi$, 
and $V_{\bar{\theta}}(s_{t+1})$ is estimated using a target network with exponentially moving parameters $\bar{\theta}$ for the online network $Q_{\theta}^{\rm core}$.
Since the action space is discrete, we can calculate  $V^{\rm core}_{\bar{\theta}}(s_{t+1})$ directly as:
\begin{eqnarray}
V^{\rm core}_{\bar{\theta}}(s_{t+1}) = {\pi_{\phi}^{\rm core}}(\cdot|s_{t+1})^{\mathrm{T}}[Q_{\bar{\theta}}^{\rm core}(\cdot|s_{t+1})-\alpha^{\rm core} \log\big(\pi_{\phi}^{\rm core}(\cdot|s_{t+1}) \big)]. \label{V(theta)}
\end{eqnarray}
For the improvement of $\pi^{\rm core}_{\phi}$, we minimize the expected KL-divergence (Equation (\ref{soft_pi_definition})) after multiplying by the temperature parameter $\alpha^{\rm core}$  and ignoring the partition function, again utilizing the property of discrete policy, we have
\begin{eqnarray}
\mathcal{L}^{\rm core}(\phi) = E_{s_t\sim D}\big[ {\pi^{\rm core}_{\phi}}(\cdot|s_t)^{\mathrm{T}}[  \alpha^{\rm core} \log(\pi^{\rm core}_{\phi}(\cdot|s_t))     -Q^{\rm core}_{\theta}(\cdot|s_t)          ]   \big]. \label{J(phi)}
\end{eqnarray}
The training of $\pi_{\phi}^{\rm core}$ and $Q_{\theta}^{\rm core}$ described above is the usual form of discrete SAC \cite{Christodoulou19SACdiscrete}.
We recommend the readers who are not familiar with discrete SAC to refer to the original paper for more details.

For the outer policy iteration (i.e., to train the PIC module $\mu^{\rm pic}(s_t, a_{t-1})$), 
we parameterize $\mu^{\rm pic}_{\varphi}(s_t, a_{t-1})$ with parameters $\varphi$, thus the parameterized mixed policy is denoted as $\pi_{\phi,\varphi}(\cdot|s_t,a_{t-1})$.
We also parameterize the $Q$-function of $\pi_{\phi,\varphi}$ as $Q_{\vartheta}(s_t, a_{t-1}, a_t)$ with parameters $\vartheta$.
To train PIC module $\mu^{\rm pic}_{\varphi}$, we use SAC algorithm with the temperature parameter $\alpha$ in the scope of the mixed policy $\pi_{\phi,\varphi}$.
Similar to Equation (\ref{J(theta)}), we optimize $\vartheta$ by conducting soft policy evaluation for the mixed policy $\pi$, i.e., minimizing the soft Bellman residual of $Q_{\vartheta}$ as:
\begin{eqnarray}
\mathcal{L}^{\rm mix}(\vartheta)=E_{(s_t,a_{t-1},a_t)\sim D}\bigg[\frac{1}{2} \big( Q_{\vartheta}(s_t,a_{t-1},a_t)- (r(s_t,a_t)+\gamma E_{s_{t+1}\sim P}[V_{\bar{\vartheta}}(s_{t+1},a_{t})])   \big)^2         \bigg],
\label{J(vartheta)}
\end{eqnarray}
and $V_{\bar{\vartheta}}$ is calculated as:
\begin{eqnarray}
&~~~~~&V_{\bar{\vartheta}}(s_{t+1},a_t) 
= {\pi}(\cdot|s_{t+1},a_t)^{\mathrm{T}}[Q_{\bar{\vartheta}}(\cdot|s_{t+1},a_t)-\alpha \log\big(\pi(\cdot|s_{t+1},a_t) \big)]  \nonumber \\
&=& { \bigg( \mu^{\rm pic}_{\psi}(s_{t+1},a_{t})\delta(a_{t})  + (1-\mu^{\rm pic}_{\psi}(s_{t+1},a_{t})) \pi^{\rm core}_{\phi}(\cdot|s_{t+1})   \bigg)     }^{\mathrm{T}}    \nonumber\\
&~& \cdot \bigg[Q_{\vartheta}(\cdot|s_{t+1},a_t)-\alpha \log\bigg( \mu^{\rm pic}_{\psi}(s_{t+1},a_{t})\delta(a_{t})  + (1-\mu^{\rm pic}_{\psi}(s_{t+1},a_{t})) \pi_{\phi}^{\rm core}(\cdot|s_{t+1}) \bigg)\bigg],
\label{equation:soft_v_calc_for_mixed_policy}
\end{eqnarray}
where $\bar\vartheta$ is the target network parameters and note the calculation of expectation and entropy is based on the mixed policy $\pi(\cdot|s_{t+1},a_t)$.

To update the PIC module parameters $\varphi$, a policy improvement step of the mixed policy $\pi_{\phi,\varphi}$ is conducted while the parameters $\phi$ (i.e., the policy core $\pi^{\rm core}_{\phi}$) is kept fixed, yielding the following objective:
\begin{eqnarray}
%\mathcal{L}^{\rm mix}(\psi)
\mathcal{L}^{\rm pic}(\psi)
&=& 
%\mathcal{L}^{\rm pic}(\psi) = 
E_{s_t\sim D}\big[ {\pi(\cdot|s_t,a_{t-1})}^{\mathrm{T}}[  \alpha \log(\pi(\cdot|s_t,a_{t-1})) - Q_{\vartheta}(\cdot|s_t,a_{t-1})          ]   \big] \nonumber \\
&=& E_{s_t\sim D}\bigg[ { \big(\mu^{\rm pic}_{\psi}(s_t,a_{t-1})\delta(a_{t-1})  + (1-\mu^{\rm pic}_{\psi}(s_t,a_{t-1})) \pi_{\phi}^{\rm core}(\cdot|s_t)  \big)   }^{T}  \nonumber \\
&& \cdot[  \alpha \log\big(\mu^{\rm pic}_{\psi}(s_t,a_{t-1})\delta(a_{t-1})  + (1-{\pi}^{\rm pic}_{\psi}(s_t,a_{t-1})) \pi_{\phi}^{\rm core}(\cdot|s_t)\big)     -Q_{\vartheta}(\cdot|s_t,a_{t-1})          ] 
%  \bigg] \nonumber \\
% &=& E_{s_t\sim D}\bigg[ { \big(\mu^{\rm pic}_{\psi}(s_t,a_{t-1})\delta(a_{t-1})   -\mu^{\rm pic}_{\psi}(s_t,a_{t-1}) \pi_{\phi}^{\rm core}(\cdot|s_t)  \big)   }^{T}  \nonumber \\
%&& \cdot[  \alpha \log\big(\mu^{\rm pic}_{\psi}(s_t,a_{t-1})\delta(a_{t-1})   -{\pi}^{\rm pic}_{\psi}(s_t,a_{t-1}) \pi_{\phi}^{\rm core}(\cdot|s_t)\big)             ]   \bigg]
\end{eqnarray}

Based on above formulations, Algorithm \ref{alg:NSAC} gives the overall algorithm of NSAC.

\begin{algorithm*}[tb]
	\caption{Nested Soft Actor Critic algorithm (NSAC)}
	\label{alg:NSAC}
	\textbf{Input:} Learning rates $\lambda_{\theta}$, $\lambda_{\vartheta}$ $\lambda_{\phi}$, $\lambda_{\varphi}$ for corresponding parameters, target network update rates $\sigma^{\rm core}$, $\sigma$ and train intervals $m^{\rm core}$, $m$
	\begin{algorithmic}[1]
		\State Initialize $\pi_{\phi}^{\rm core}$, $Q^{\rm core}_{\theta_1}$, $Q^{\rm core}_{\theta_2}$, $Q^{\rm core}_{\bar{\theta_1}}$, $Q^{\rm core}_{\bar{\theta_2}}$  
		\hfill \textcolor{blue}{\# Initialize networks for policy core $\pi^{\rm core}$ %$\pi^{\rm core}(\cdot|s_t)$
		}
		\State Initialize $\mu_{\varphi}^{\rm pic}$
		\hfill \textcolor{blue}{\# Initialize Policy Inertia Controller $\mu^{\rm pic}$}
		\State Initialize $Q_{\vartheta_1}$, $Q_{\vartheta_2}$, 
		%$\xi_{\psi} \geq \xi_0$,  
		$Q_{\bar{\vartheta_1}}$, $Q_{\bar{\vartheta_2}}$
		\hfill \textcolor{blue}{\# Initialize networks for mixed policy $\pi$ %$\pi(\cdot|s_t,a_{t-1}$
		}
		\State Initialize $D \leftarrow \emptyset $
		\hfill \textcolor{blue}{\# Initialize replay buffer}
		\State Set $\bar{\theta_1} \leftarrow \theta_1 $, $\bar{\theta_2} \leftarrow \theta_2 $, $\bar{\vartheta_1} \leftarrow \vartheta_1 $, $\bar{\vartheta_2} \leftarrow \vartheta_2 $
		\hfill \textcolor{blue}{\# Equalize target and local network weights}
		\For{each environment step $t$ from $0$ to $T$}	
		    \State $a_t \sim \pi(s_t,a_{t-1})$ 
		    \hfill \textcolor{blue}{\# Sample action from the mixed policy $\pi$. Specially, $a_{-1}$ is set to be a default/null action}
		    \State $s_{t+1} \sim P(s_{t+1}|s_t,a_t)$,   
		    \hfill \textcolor{blue}{\# Sample transition from the environment}
		    \State $D \leftarrow D \cup \{ \big( s_t, a_{t-1}, a_t, r(s_t,a_t) ,s_{t+1} \big) \}$   
		    \hfill \textcolor{blue}{\# Store the transition in the replay buffer}
		    \If{step $t \bmod m^{\rm core} \equiv 0$}
    		    \State $\theta_{i} \leftarrow \theta_{i} - \lambda_{\theta} \nabla_{\theta_i}{\mathcal{L}^{\rm core}(\theta_i)} $ for $i \in {1,2} $ 
    		    \hfill \textcolor{blue}{\# Update the parameters of $Q^{\rm  core}$}
    	        \State $\phi \leftarrow \phi - \lambda_{\phi}\nabla_{\phi}{\mathcal{L}^{\rm core}(\phi)} $ \hfill \textcolor{blue}{\# Update the parameters of $\pi^{\rm core}(\cdot|s_t)$}
    	        \State $\bar{\theta}_i \leftarrow \sigma^{\rm core}\theta_i +(1-\sigma^{\rm core})\bar{\theta}_i $ for $i={1,2}$ 
    	        \hfill \textcolor{blue}{\# Update the parameters of target $Q^{\rm core}$}
	        \EndIf
    	    \If{step $t \bmod m \equiv 0$}
    	        \State $\vartheta_{i} \leftarrow \vartheta_{i} - \lambda_{\vartheta} \nabla_{\vartheta_i}{\mathcal{L}^{\rm mix}(\vartheta_i)} $ for $i \in {1,2} $
    	        \hfill \textcolor{blue}{\# Update the parameters of $Q$}
    	        \State $\psi \leftarrow \psi - \lambda_{\varphi}\nabla_{\psi}{\mathcal{L}^{\rm pic}(\psi)} $  \hfill \textcolor{blue}{\# Update the parameters of $\mu^{\rm pic}(s_t,a_{t-1})$}
    	        \State $\bar{\vartheta}_i \leftarrow \sigma\vartheta_i +(1-\sigma)\bar{\vartheta}_i $ for $i={1,2}$
    	        \hfill \textcolor{blue}{\# Update the parameters of target $Q$}
    		\EndIf
		\EndFor
		\State Output $\theta_1$, $\theta_2$, $\phi$, $\vartheta_1$, $\vartheta_2$, $\psi$   
		\hfill \textcolor{blue}{\# Output the optimized parameters }
	\end{algorithmic}
\end{algorithm*}

\section{Experimental Details}
\label{section:exp_details}

\subsection{Environment Setup}
We conduct our experiments on \emph{Highway} autonomous driving simulators, which is provided at \url{https://github.com/eleurent/highway-env}.
A few modifications are maded to make the rendering of the environments faster.
For the detailed configurations for the selected four environments, please see Section \ref{section:highway_config}.
Besides, we also adopt several representative tasks in OpenAI Atari games, i.e., \emph{MsPacman}, \emph{SpaceInvaders}, \emph{Qbert}, \emph{JamesBond}.
More details can be found at \url{https://gym.openai.com/envs/#atari}.

\subsection{Configurations of \emph{Highway} Tasks}
\label{section:highway_config}
We follow the built-in configurations in \emph{Highway} as originated presented at \url{https://github.com/eleurent/highway-env},
and add some extra configurations in Table \ref{table:configuration_in_highway}.
\begin{table}[ht]
  \caption{Detail configurations of four \emph{Highway} tasks. [$a$, $b$] denotes a Uniform distribution on the range from integer $a$ to integer $b$.}
  \label{table:configuration_in_highway}
  \centering
  \scalebox{1.2}{
  \begin{tabular}{ccccc}
    \toprule
    Configuration Items & \emph{Lane Change} & \emph{Merge} & \emph{Intersection} 
      & \emph{Two-Way} \\
    \midrule
    Vehicles Count for Training &  $[20,50]$ & $[6,12]$  & $10$ & $10$ \\
    Vehicles Count for Testing & $45$ &   $[6,12]$ & $10$ & $10$\\
    Vehicles Count in Ego Vehicle's Scope & $10$ & $10$ & $5$ & $10$\\
    \midrule
    COLLISION REWARD & $-1$ & $-1$ & $0$ & $0$\\
    RIGHT LANE REWARD & $0$ & $0.1$ & $0$ & $0$\\
    HIGH VELOCITY REWARD &$0.4$ &$0.2$ &$0$ & $0.8$\\
    LANE CHANGE REWARD &$-0.1$ & $-0.05$& $0$ & $0$\\
    MERGING VELOCITY REWARD & $0$ & $-0.5$ & $0$ & $0$\\
    ARRIVED REWARD &$0$ & $0$ & $5$ & $0$ \\
    LEFT LANE CONSTRAINT & $0$ & $0$ & $0$ & $1$\\
    LEFT LANE REWARD  & $0$ & $0$ & $0$ & $0.2$ \\ 
    \midrule
    Duration & $70$s & $25$s & $25$s & $25$s\\
    \bottomrule
  \end{tabular}
  }
\end{table}

\subsection{Implementation Details}
\label{section:implementation_details}
Our codes are implemented with Python 3.6 and Tensorflow.
For discrete Soft Actor-Critic (SAC) algorithm, we use a modified version of the implementation at \url{https://github.com/ku2482/sac-discrete.pytorch} to suit our environments,
and based on which we implement our algorithm Nested Soft Actor-Critic (NSAC).
For Deep Q-Network (DQN), we use a standard implementation of the vanilla version, i.e., with only experience replay and target network. For the action repetition approach, we set the repetition set as $\mathit{Re}=\{1,2,4,8\}$ , which means that the action is repeated for $1,2,4,8$ times, then the augmented action space $\mathit{A}^{\prime}$ is the Cartesian product $\mathit{A} \times \mathit{Re}$. \emph{DQN-repeat} baseline and \emph{SAC-repeat} baseline mean that DQN and SAC are trained on the augmented action space $\mathit{A}^{\prime}$, respectively. The hyperparameter settings for \emph{DQN-repeat} baseline and \emph{SAC-repeat} baseline are the same as DQN and SAC.
Our codes will be released on Github soon.
% if our paper is accepted.
%For more code-level details, please see our codes in the supplementary material.

\textbf{Network Structures.}
The network structures of DQN/SAC/NSAC used for \emph{Highway} autonomous driving environments and OpenAI Atari games are shown in Table \ref{table:networks_in_highway} and Table \ref{table:networks_in_atari} respectively.
We use a two-layer fully-connected neural network in \emph{Highway} and an additional convolutional neural network is adopted in Atari games for pixel inputs.

\begin{table}[ht]
  \caption{Network structures of approaches in \emph{Highway} autonomous driving environments.
  }
  \label{table:networks_in_highway}
  \centering
  \scalebox{1.0}{
  \begin{tabular}{cccc}
    \toprule
    Layer & Actor Network & Critic or $Q$ Network & PIC Network \\
      & ($\pi^{\rm core}(s_t)$) & ($Q^{\rm core}(s_t,a_t)$ and $Q(s_t,a_{t-1},a_t)$) & ($\mu^{\rm pic}(s_t, a_{t-1})$)\\
    \midrule
    Fully-connected & (state dim, 64) & (state dim, 64) or (state dim + action dim, 64) & (state dim + action dim, 64) \\
    Activation & ReLU & ReLU & ReLU\\
    \midrule
    Fully-connected & (64, 64) & (64, 64) & (64, 64) \\
    Activation & ReLU & ReLU & ReLU \\
    \midrule
    Fully-connected & (64, action dim) & (64, action dim) & (64, 1) \\
    Activation & softmax & None & tanh \\
    \bottomrule
  \end{tabular}
  }
\end{table}

\begin{table}[ht]
  \caption{Network structures of approaches in OpenAI Atari games.
  }
  \label{table:networks_in_atari}
  \centering
  \scalebox{1.1}{
  \begin{tabular}{ccc}
    \toprule
    Network & Layer (Name) & Structure \\
    \midrule
    Convolutional Network & Convolution & 32 channels, 8x8 kernel, 4 stride, 0 padding \\
     & Activation & ReLU \\
     & Convolution & 64 channels, 4x4 kernel, 2 stride, 0 padding \\
     & Activation & ReLU \\
     & Convolution & 64 channels, 3x3 kernel, 1 stride, 0 padding \\
     & Activation & ReLU \\
     & Flatten & (into size [batch size, 7*7*64])\\
    \midrule
    Fully-connected Network & Fully-connected & (7*7*64, 512) for $\pi^{\rm core}$ and $Q^{\rm core}$;  \\
     &  & (7*7*64 + action dim, 512) for $\mu^{\rm pic}$ and $Q$  \\
     & Activation & ReLU \\
     & Fully-connected & (512, action dim) for $\pi^{\rm core}$, $Q^{\rm core}$ and $Q$; \\
     &  & (512, 1) for $\mu^{\rm pic}$  \\
     & Activation & softmax for $\pi^{\rm core}$; None for $Q^{\rm core}$ $Q$; tanh for $\mu^{\rm pic}$ \\
    \bottomrule
  \end{tabular}
  }
\end{table}

% \textbf{Hyperparameters.}
% Table \ref{table:hyperparameter} shows the common hyperparamters of algorithms used in our experiments.
% No regularization is used for the actor and the critic in all algorithms.

\begin{table}[ht]
  \caption{
  Hyperparameter setting of algorithms adopted.
  %We use `-' to denote the `not applicable' situation.
  }
  \label{table:hyperparameter}
  \centering
  \scalebox{1.2}{
  \begin{tabular}{c|c|c}
    \toprule
    Hyperparameter & \emph{Highway} & Atari\\
    \midrule
    Learning Rate & 3$\cdot$10$^{-4}$ & 3$\cdot$10$^{-4}$ \\
    Discount Factor & 0.99 & 0.99 \\
    Batch Size & 64 & 64 \\
    Buffer Size & $2\cdot 10^{5}$ & 10$^{6}$ \\
    Optimizer & Adam & Adam \\
    \midrule
    Policy Inertia Controller Learning Rate (for NSAC) & 3$\cdot$10$^{-4}$ & 3$\cdot$10$^{-3}$ \\
    Soft Target Replacement Rate (for SAC/NSAC) & $0.002$ & $0.005$ \\
    Hard Target Replacement Rate (for DQN) & 10$^{4}$ & 10$^{4}$ \\
    Temperature Parameter (for SAC and $\pi^{\rm core}$ in NSAC) & 0.1 & 0.1 \\
    Temperature Parameter (for mixed policy $\pi$ in NSAC) & 0.01 & 0.0001 \\
    \bottomrule
  \end{tabular}
  }
\end{table}

\textbf{Training Details.}
%For \emph{Highway} autonomous driving environments, 
%we use the original states and rewards, no additional normalization is used.
For all approaches, we update the parameters every 2 transition samples are collected in \emph{Highway} autonomous driving environments, and every 4 transition samples for Atari games.
%, we update the parameters every 4 transition samples are collected.
For DQN and its variants, $\epsilon$-greedy is adopted for exploration. 
We gradually schedule $\epsilon$ from 1 to 0.1 through decaying it with a decrease of 5 $\cdot$ 10$^{-6}$ for each update for both \emph{Highway} tasks and Atari games. 
For NSAC, the policy core $\pi^{\rm core}$ (inner policy iteration) and the PIC module $\mu^{\rm pic}$ (outer policy iteration) is updated at a $1:1$ frequency.
Moreover, we do not use the code-level tricks, such as state normalization, reward normalization, gradient clip and etc.

\subsection{Hyperparameters}
Table \ref{table:hyperparameter} shows the common hyperparamters of algorithms used in our experiments.
No regularization is used for the actor and the critic in all algorithms.

\end{document}